\documentclass[10pt,twocolumn,letterpaper]{article}

\usepackage[pagenumbers]{cvpr} %

\usepackage[textsize=tiny]{todonotes}
\usepackage{color, colortbl}
\usepackage{xcolor}
\usepackage{pgf} %
\definecolor{Gray}{gray}{0.9}
\usepackage{tikz} %
\usepackage{pgffor} %

\usepackage{graphicx} %

\usepackage{tikz, pgffor} %

\usepackage[table]{xcolor}
\usepackage{pgfplotstable}
\usepackage{graphicx} %
\usepackage{float} %
\usepackage{booktabs} %
\usepackage{amsmath}
\usepackage{multirow}
\usepackage{fontawesome5}

\usepackage[ruled]{algorithm2e}
\usepackage{mathtools}

\pgfplotstableset{
    color cells/.style={
        col sep=comma,
        string type,
        postproc cell content/.code={%
                \pgfkeysalso{@cell content=\rule{0cm}{2.4ex}\cellcolor{black!##1}\pgfmathtruncatemacro\number{##1}\ifnum\number>50\color{white}\fi##1}%
                },
        columns/x/.style={
            column name={},
            postproc cell content/.code={}
        }
    }
}

\definecolor{cvprblue}{rgb}{0.21,0.49,0.74}
\usepackage[pagebackref,breaklinks,colorlinks,allcolors=cvprblue]{hyperref}

\def\paperID{2536} %
\def\confName{CVPR}
\def\confYear{2026}

\title{MAPS: Preserving Vision-Language Representations via Module-Wise Proximity Scheduling for Better Vision-Language-Action Generalization}

\author{Chengyue Huang\textsuperscript{*} \quad Mellon M. Zhang\textsuperscript{*} \quad Robert Azarcon \\ Glen Chou \quad Zsolt Kira\\
Georgia Institute of Technology\\
{\tt\small \{chuang475,meilongz,razarcon3,chou,zkira\}@gatech.edu}\\
\href{https://mapsvla.github.io/}{\tt \small mapsvla.github.io}
}

\begin{document}
\maketitle
\def\thefootnote{*}\footnotetext{Equal contribution.}
\begin{abstract}
Vision-Language-Action (VLA) models inherit strong priors from pretrained Vision-Language Models (VLMs), but naïve fine-tuning often disrupts these representations and harms generalization. Existing fixes -- freezing modules or applying uniform regularization -- either overconstrain adaptation or ignore the differing roles of VLA components. We present \textbf{MAPS (Module-Wise Proximity Scheduling)}, the first robust fine-tuning framework for VLAs. Through systematic analysis, we uncover an empirical order in which proximity constraints should be relaxed to balance stability and flexibility. MAPS linearly schedules this relaxation, enabling visual encoders to stay close to their pretrained priors while action-oriented language layers adapt more freely. MAPS introduces no additional parameters or data, and can be seamlessly integrated into existing VLAs. Across MiniVLA-VQ, MiniVLA-OFT, OpenVLA-OFT, and challenging benchmarks such as SimplerEnv, CALVIN, LIBERO, as well as real-world evaluations on the Franka Emika Panda platform, MAPS consistently boosts both in-distribution and out-of-distribution performance (up to +30\%). Our findings highlight empirically guided proximity to pretrained VLMs as a simple yet powerful principle for preserving broad generalization in VLM-to-VLA transfer.
\end{abstract}    
\vspace{-1em}
\section{Introduction}
\label{sec:intro}

Vision-Language-Action (VLA) models have recently emerged as a unifying framework for robotic learning, enabling agents to perceive their environment, understand natural language instructions, and execute actions in an end-to-end manner \cite{kim2024openvlaopensourcevisionlanguageactionmodel, brohan2023rt1roboticstransformerrealworld, brohan2023rt2visionlanguageactionmodelstransfer, black2024pi0visionlanguageactionflowmodel}. However, the scarcity of large-scale robotics data makes it infeasible to train these models from scratch. As a result, most VLA systems initialize from powerful vision-language models (VLMs) \cite{beyer2024paligemmaversatile3bvlm, bai2025qwen25vltechnicalreport, touvron2023llama2openfoundation} pretrained on web-scale data, leveraging their rich visual and semantic priors before finetuning on robotics datasets to learn action policies. However, while this strategy has led to impressive task performance, it often comes at the cost of generalization -- models degrade in performance when faced with novel tasks, objects, or environments \cite{dey2024revlarevertingvisualdomain,grover2025enhancinggeneralizationvisionlanguageactionmodels}.

This loss of generalization arises from a fundamental mismatch between large-scale pretraining and the limited scope of robot-specific finetuning \cite{kim2024openvlaopensourcevisionlanguageactionmodel}. Robotics datasets are typically sparse and task-specific, and finetuning extensively on them induces spurious correlations and overfitting, which in turn causes catastrophic forgetting 
of spatial reasoning, world knowledge, and linguistic grounding.
The core challenge in VLA training therefore lies in balancing the tradeoff between action adaptation and preservation of pretrained generalization. Excessive finetuning causes the model to drift too far from its pretrained initialization, diminishing robustness to distribution shifts \cite{tian2023trainableprojectedgradientmethod}.
Conversely, insufficient finetuning limits the model’s ability to align with the action space.

Several approaches attempt to balance adaptation and generalization. Freezing visual encoders preserves pretrained perception but limits adaptation to robotics-specific cues. Dual-encoder designs \cite{grover2025enhancinggeneralizationvisionlanguageactionmodels} maintain both frozen and trainable visual pathways, improving flexibility at the cost of doubled memory and compute. Weight interpolation techniques \cite{dey2024revlarevertingvisualdomain} gradually revert visual weights to their pretrained state while adapting the language backbone, but require multiple training stages. In these efforts, generalization often comes at the expense of higher computational and architectural complexity, with most methods emphasizing visual preservation while overlooking the role of semantic priors in language components.

Seeking a lightweight alternative to preserving pretrained knowledge, we begin by revisiting the common practice of model freezing. Through a systematic analysis of freezing different VLM components during action finetuning, we confirm a widely held but rarely quantified intuition: visual modules should remain strongly constrained. Early visual backbones such as DINOv2 \cite{oquab2024dinov2learningrobustvisual} capture essential geometric and spatial priors for manipulation, whereas language-aligned modules like SigLIP \cite{zhai2023sigmoidlosslanguageimage} and deeper language layers benefit from greater flexibility to adapt task-specific semantics and action grounding. However, we also find that freezing itself introduces task-specific inductive biases, and certain freeze configurations fail to generalize across new domains and environments.

Instead of hard constraints from model freezing, soft regularization offers a more flexible and robust alternative with comparable efficiency \cite{tian2024rethinkingweightdecayrobust, li2018explicitinductivebiastransfer, tian2023trainableprojectedgradientmethod, tian2023fasttrainableprojectionrobust, gouk2020regularisationneuralnetworksenforcing}. Projection-based selective tuning methods, such as selective projection decay \cite{tian2024rethinkingweightdecayrobust}, replace conventional weight decay with a projection operator that limits how far parameters drift from their pretrained values -- enabling adaptive regularization across layers without added complexity. However, existing approaches typically use a single hyperparameter to enforce a uniform regularization strength, implicitly assuming all layers should deviate equally. We find this assumption ill-suited for VLA models, where different components encode distinct priors and vary in their sensitivity to fine-tuning. Some modules must stay near their pretrained state to preserve structural knowledge, while others require greater flexibility to adapt task-specific grounding.

This introduces a new tradeoff: naive freezing provides modularity but is brittle due to strong inductive bias, whereas robust finetuning relaxes this bias but sacrifices modularity. To bridge this tradeoff and successfully adapt the robust finetuning paradigm to the action domain, we propose \textbf{MAPS (Module-Wise Proximity Scheduling)}, a robust finetuning framework designed specifically for VLA models. MAPS extends traditional projection-based selective tuning \cite{tian2024rethinkingweightdecayrobust} by modulating the 
\textit{projection strength rate of change} hyperparameter across model components and training steps via a linear scheduler. This encourages more careful updates of the visual components to preserve pretrained geometric priors while allowing more explorative updates of the language components to quickly adapt to the action space. Importantly, this scheduling is implemented with no additional models, data, or parameters and integrates seamlessly with existing VLA architectures.

We validate MAPS across a wide range of VLA backbones, including MiniVLA-VQ \cite{belkhale2024minivla, lee2024behaviorgenerationlatentactions}, MiniVLA-OFT, OpenVLA-OFT \cite{kim2025finetuningvisionlanguageactionmodelsoptimizing}, etc. Experiments on a diverse set of evaluation benchmarks - SimplerEnv Bridge Data \cite{li2024evaluatingrealworldrobotmanipulation}, CALVIN \cite{mees2022calvinbenchmarklanguageconditionedpolicy} (ABC→D), LIBERO \cite{liu2023liberobenchmarkingknowledgetransfer} (train on LIBERO-90, test on Goal/Spatial/Object/Long), and real-world evaluations on a Franka Emika Panda robot consistently show that MAPS improves out-of-distribution (OOD) generalization while preserving pretrained structure and in-distribution (ID) performance. These findings suggest that maintaining proximity to pretrained VLMs in a selective manner is a simple yet powerful principle that could guide future scaling and pre-training of VLA systems.
\section{Related Works}
\label{sec:related_work}

\noindent \textbf{Vision-Language-Action Models.} 
Vision-Language-Action (VLA) models advance unified embodied intelligence by leveraging large pretrained foundation models~\cite{beyer2024paligemmaversatile3bvlm, bai2025qwen25vltechnicalreport, touvron2023llama2openfoundation} and their components~\cite{oquab2024dinov2learningrobustvisual, zhai2023sigmoidlosslanguageimage, radford2021learningtransferablevisualmodels}. Systems like RT-1, PaLM-E, and RT-2~\cite{brohan2023rt1roboticstransformerrealworld, driess2023palmeembodiedmultimodallanguage, brohan2023rt2visionlanguageactionmodelstransfer} showed that grounding vision-language pretraining in experience enables cross-task transfer. Large frameworks such as Open-X Embodiment, OpenVLA, and $\pi_0$~\cite{embodimentcollaboration2025openxembodimentroboticlearning, kim2024openvlaopensourcevisionlanguageactionmodel, kim2025finetuningvisionlanguageactionmodelsoptimizing, black2024pi0visionlanguageactionflowmodel, intelligence2025pi05visionlanguageactionmodelopenworld, driess2025knowledgeinsulatingvisionlanguageactionmodels} scale data and architectures for generalization. Recent work improves action tokenization~\cite{szot2024grounding, pertsch2025fastefficientactiontokenization, belkhale2024rthactionhierarchiesusing, lee2024behaviorgenerationlatentactions, zhao2023learningfinegrainedbimanualmanipulation, reuss2025flower}, efficiency~\cite{shukor2025smolvlavisionlanguageactionmodelaffordable, wen2025tinyvlafastdataefficientvisionlanguageaction, belkhale2024minivla}, and grounding/reasoning~\cite{zhou2025chatvlaunifiedmultimodalunderstanding, zhou2025chatvla2visionlanguageactionmodelopenworld, shi2025hirobotopenendedinstruction, lee2025molmoactactionreasoningmodels, qu2025spatialvlaexploringspatialrepresentations, tian2024predictiveinversedynamicsmodels, szot2024multimodalllmsgeneralistembodied}, while emerging work explores chain-of-thought reasoning for interpretable control~\cite{zhao2025cotvlavisualchainofthoughtreasoning, zawalski2025roboticcontrolembodiedchainofthought, wen2025diffusionvlageneralizableinterpretablerobot}. Despite progress, most VLAs rely on fine-tuning that disrupts pretrained priors; MAPS instead preserves them via module-wise proximity scheduling for selective adaptation.

\noindent \textbf{VLA Generalization.}
Improving generalization in VLA models focuses on preserving pretrained representations while adapting to new tasks. Data-oriented approaches reformulate robotic data for better VLM alignment (directional prediction~\citep{zhang2025inspirevisionlanguageactionmodelsintrinsic} or textual trajectory conversion~\citep{hancock2025actionslanguagefinetuningvlms}) and co-train with action objectives~\citep{grover2025enhancinggeneralizationvisionlanguageactionmodels, szot2024grounding}. Training-oriented methods retain learned priors via weight interpolation~\citep{dey2024revlarevertingvisualdomain}, dual encoders~\citep{grover2025enhancinggeneralizationvisionlanguageactionmodels}, frozen cross-attention~\citep{huang2025ottervisionlanguageactionmodeltextaware}, embedding regularization~\citep{kachaev2025dontblindvlaaligning}, or action perturbation~\citep{guo2025robustnessvisionlanguageactionmodelmultimodal}. These add computational overhead, whereas MAPS constrains updates by module importance and proximity to pretrained weights, preserving priors and improving OOD generalization without extra data or parameters.

\begin{figure*}[!h]
    \centering
    \includegraphics[width=\linewidth]{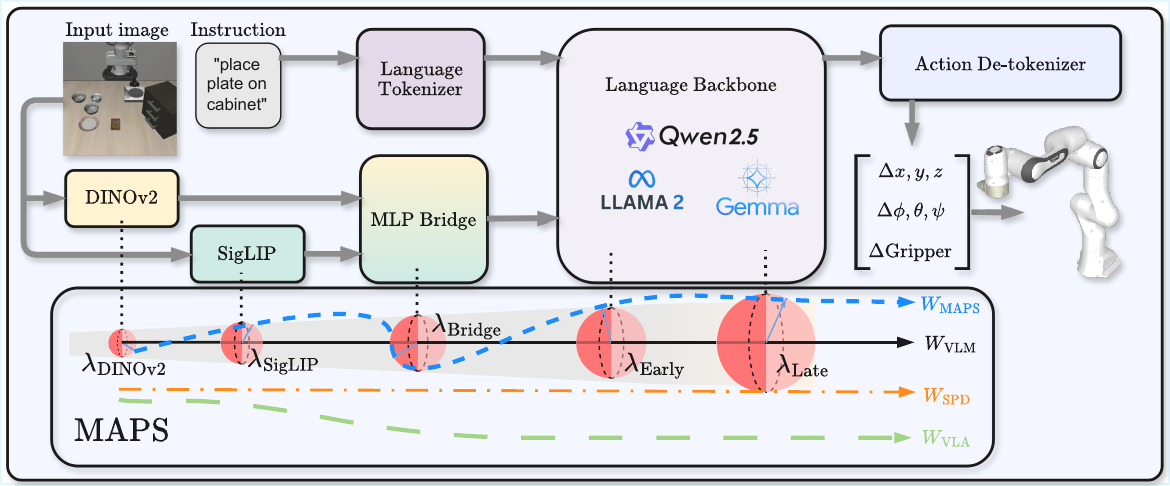}
    \caption{\textbf{Module-Wise Proximity Scheduling (MAPS)}. MAPS is applied on pretrained VLM components during the action finetuning stage. MAPS (blue dash line) enforces strong preservation on early vision layers while progressively relaxing constraints toward higher-level language layers. In contrast, vanilla finetuning (green dash line) distorts the VLM representation completely away from its pretrained weights (black solid line), and uniform SPD (orange dash/dot line) applies the same constraint everywhere.} 
    \label{fig:maps}
    \vspace{-1em}
\end{figure*}
\section{Background}
\label{sec:prelim}

\subsection{Vision-Language-Action Models}

A Vision-Language-Action (VLA) model \cite{kim2024openvlaopensourcevisionlanguageactionmodel,black2024pi0visionlanguageactionflowmodel,tian2023fasttrainableprojectionrobust,brohan2023rt2visionlanguageactionmodelstransfer} seeks to learn a policy $\pi_\theta(a_t \mid o_t, x)$ that maps a multimodal observation $o_t$ and a language instruction $x$ to an action $a_t$ in an embodied environment. Typically, the visual and language inputs are encoded by $f_v$ and $f_l$ separately, and fused by a multimodal transformer $f_m$ to produce a policy embedding $z_t = f_m([f_v(o_t), f_l(x)])$. Action parameterization varies across families of VLAs: \textbf{continuous-action models} predict low-level control vectors directly and are trained to make their outputs closely match the expert’s actions, using objectives such as $\ell_1$ regression \cite{kim2025finetuningvisionlanguageactionmodelsoptimizing}, diffusion-based denoising \cite{wen2025diffusionvlageneralizableinterpretablerobot}, or flow matching \cite{black2024pi0visionlanguageactionflowmodel}. In contrast, \textbf{discrete-action models} convert actions or short action segments into a finite set of tokens \cite{kim2024openvlaopensourcevisionlanguageactionmodel}.
Modern VLAs build upon pretrained vision-language backbones such as \textsc{DINOv2} \cite{oquab2024dinov2learningrobustvisual}, \textsc{SigLIP} \cite{zhai2023sigmoidlosslanguageimage}, and \textsc{Llama} \cite{touvron2023llama2openfoundation}, using either a multi-modal transformer or a lightweight cross-attention ``bridge'' to fuse representations before an action head maps $z_t$ into either continuous controls or discrete tokens. Models such as $\pi_0$-Fast \cite{pertsch2025fastefficientactiontokenization}, and OpenVLA-OFT \cite{kim2025finetuningvisionlanguageactionmodelsoptimizing} follow this template but introduce small projection heads or adapters for rapid finetuning and deploying. During training on robot demonstrations, the model grounds linguistic semantics in action-conditioned representations that transfer across tasks, objects, and embodiments.

\subsection{Robust Fine-Tuning}
\label{sec:spd}

We revisit projection-based robust fine-tuning (RFT) methods \cite{tian2023trainableprojectedgradientmethod,tian2023fasttrainableprojectionrobust,tian2024rethinkingweightdecayrobust,huang2025directionalgradientprojectionrobust,huang2025framesvqabenchmarkingfinetuningrobustness} that constrain fine-tuning to remain close to the pretrained weights, enabling models to adapt to downstream tasks while preserving pretrain generalization.

\noindent\textbf{From L2 regularization to projection.}
Vanilla fine-tuning updates model parameters to minimize the task loss $\mathcal{L}$ without constraining deviation from the pretrained model. This often leads to feature drift and catastrophic forgetting \cite{li2018explicitinductivebiastransfer,tian2023trainableprojectedgradientmethod,tian2023fasttrainableprojectionrobust}. L2-SP \cite{li2018explicitinductivebiastransfer} addresses this issue by penalizing the distance between the fine-tuned parameters $\theta_t$ and the pretrained initialization $\theta_0$ with regularization strength $\lambda_\text{reg}$ at each gradient iteration step $t$:
\begin{equation}
    \mathcal{L}_{\text{L2-SP}} = \mathcal{L}(\theta_t) + \frac{\lambda_{\text{reg}}}{2}\|\theta_t - \theta_0\|_2^2.
\end{equation}
Instead of regularizing $\|\theta_t\|_2$, L2-SP explicitly preserves proximity to the pretrained model. 
Building on this idea, TPGM \cite{tian2023trainableprojectedgradientmethod} reformulates L2-SP by enforcing the model to stay within a distance of $\gamma$ from the pre-trained model:
\begin{align}
\label{eq:tpgm}
    \min \; \mathcal{L}(x,y;\theta_t), \quad 
    \text{s.t.} \quad \|\theta_t - \theta_0\|_2 \leq \gamma.
\end{align}
To enforce this constraint during training, TPGM applies the \textit{Projected Gradient Method} (PGM), projecting the updated weights back into an $\ell_2$ ball with radius $\gamma$ around $\theta_0$, where $\tilde\theta_t$ is the unconstrained updated weight:
\begin{align}
\label{eq:l2_proj}
    \theta_t
    = \theta_0 + 
    \frac{1}{\max\left(1, \frac{\|\tilde{\theta}_t - \theta_0\|_2}{\gamma} \right)} 
    (\tilde{\theta}_t - \theta_0).
\end{align}

\noindent\textbf{Selective Projection Decay (SPD).}
SPD \citep{tian2024rethinkingweightdecayrobust} builds on the projection view of L2-SP and TPGM by determining whether and how much the uncontrained update $\tilde{\theta}_t$ should be projected back toward the pretrained weights $\theta_0$.

\noindent\textbf{Step 1: When to project.}  
Motivated by the derivative of $\mathcal{L}$ w.r.t. $\lambda_\text{reg}$, i.e. $\frac{\partial \mathcal{L}(\theta_{t})}{\partial\lambda_\text{reg}} = \frac{\partial \mathcal{L}(\theta_{t})}{\partial \theta_{t}}^\intercal\frac{\partial \theta_{t}}{\partial \lambda_\text{reg}} = \alpha*-g_{t+1}^\intercal(\Tilde{\theta}_t-\theta_0)$ where $\alpha$ is the learning rate and $g$ is the gradient, SPD uses the sign of $c_t := -g_{t+1}^\top(\tilde{\theta}_t - \theta_0)$ as the selection condition, since it determines the direction of the hyper-update: if $c_t < 0$, gradient descent increases $\lambda_\text{reg}$ and strengthens L2-SP regularization, whereas if $c_t > 0$, it decreases $\lambda_\text{reg}$ and weakens the regularization.
Thus,
SPD triggers a projection step if $c_t < 0$; otherwise the update is left unchanged.

\noindent\textbf{Step 2: How much to project.}  
L2-SP's optimization step after unconstrained update can be written as ${\theta}_t = \Tilde{\theta}_t - \lambda_\text{reg}\alpha(\Tilde{\theta}_t-\theta_0)$. Reformulating \cref{eq:l2_proj} as $\theta_t = \Tilde{\theta}_t - \left(1-\frac{\gamma}{\max\{ \gamma,\|\Tilde{\theta}_t-\theta_0\|_2\}}\right) * (\Tilde{\theta}_t - \theta_0)$, L2-SP (regularization) and projection are equivalent if $\lambda_\text{reg}\alpha = (1-\frac{\gamma}{\max\{ \gamma,\|\Tilde{\theta}_t-\theta_0\|_2\}})$. SPD then defines a deviation ratio $r_t=\frac{\max \{ \gamma_t - \gamma_{t-1}\}}{\gamma_t}$ based on the current deviation radius (before regularization) $\gamma_t = \|\tilde{\theta}_t - \theta_0\|_2$ and the previous radius $\gamma_{t-1} = \|\theta_{t-1} - \theta_0\|_2$, and replace the learning rate $\alpha$ in L2-SP with the deviation ratio $r_t$. This makes the \textit{new hyper-parameter $\lambda$} tuning more intuitive, where $\lambda_\text{reg}\alpha=\lambda r_t$.
Intuitively, with $\lambda=1$, 
$\theta_t = \Tilde{\theta}_t - \frac{\max\{0,\gamma_t - \gamma_{t-1}\}}{\gamma_t}(\Tilde{\theta}_t-\theta_0) = \theta_0 + \frac{\gamma_{t-1}}{\max\{ \gamma_{t-1},\gamma_t\}} * (\Tilde{\theta}_t - \theta_0)$, the regularization is equivalent to projecting the updated parameters onto a ball with radius equal to the previous deviation whenever the current deviation exceeds that value.
Now, a single hyperparameter $\lambda$ replaces the conventional weight decay and controls how aggressively the constraint radius expands or contracts during training. 
When $\lambda = 0$, SPD reduces to full fine-tuning with no projection; $0 < \lambda < 1$ allows the radius to expand, enabling flexible adaptation; $\lambda = 1$ keeps the deviation from the pretrained model fixed; and $\lambda > 1$ contracts the radius, pulling parameters back toward $\theta_0$. Despite being a single scalar, $\lambda$ enables dynamic and layer-specific projection across optimization steps.
\section{Method}
\label{sec:method}

\begin{figure}[!t]
    \centering

    \begin{minipage}{0.24\linewidth}
        \centering
        \includegraphics[width=\linewidth]{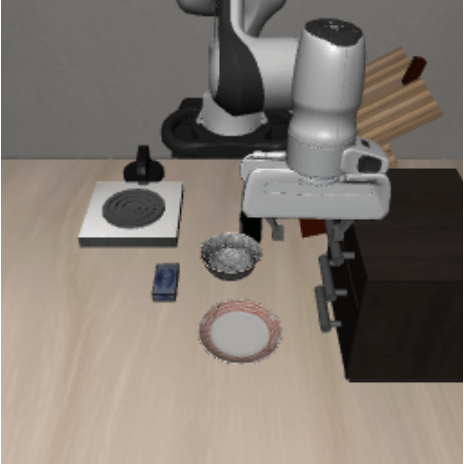}
    \end{minipage}
    \begin{minipage}{0.24\linewidth}
        \centering
        \includegraphics[width=\linewidth]{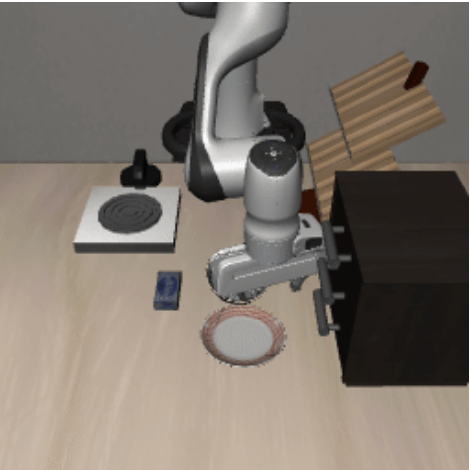}
    \end{minipage}
    \begin{minipage}{0.24\linewidth}
        \centering
        \includegraphics[width=\linewidth]{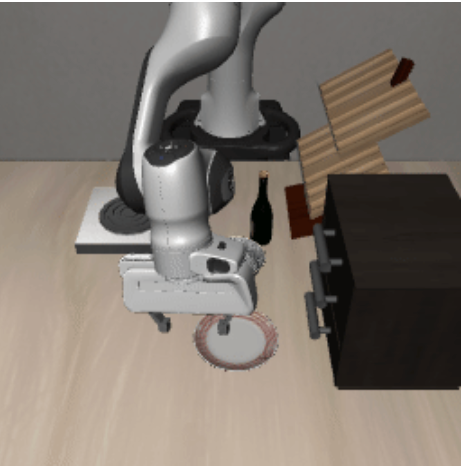}
    \end{minipage}
    \begin{minipage}{0.24\linewidth}
        \centering
        \includegraphics[width=\linewidth]{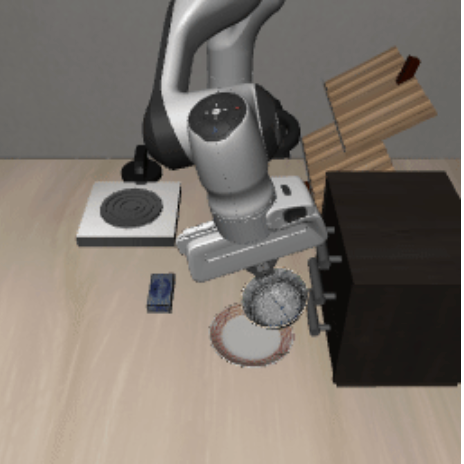}
    \end{minipage}

    \caption{Qualitative comparison of freezing configurations for LIBERO-90 task "put the bowl on the plate" (left to right: full fine-tuning (FFT), freeze VLM, freeze language, freeze vision). FFT fails most, targetting the cabinet instead of the bowl. Freezing VLM/language/vision preserves 2D localization of the bowl but impairs depth reasoning. When vision is frozen, the policy can grasp the bowl but fails to accurately place it on the plate.}
    \vspace{-1em}
    \label{fig:prelim_viz}
\end{figure}

\begin{table*}[!ht]
\centering
\caption{Comparison of freezing or robust fine-tuning (RFT) different parts of VLA on SimplerEnv's ID and OOD tasks (details in \cref{sec:main_exp_simpler}).  Early L and Last L denote early and the last language layers. \textcolor{cyan}{\faSnowflake} and \textcolor{red}{\faFire} represent freeze and full fine-tune.}
\vspace{-1em}
\resizebox{1.0\linewidth}{!}{%
\begin{tabular}{cccc|ccccc|cccc}
\toprule
\multicolumn{4}{c|}{\textbf{Modules}} & \multicolumn{5}{c|}{\textbf{ID}} & \multicolumn{4}{c}{\textbf{OOD}} \\
\cmidrule(lr){1-4} \cmidrule(lr){5-9} \cmidrule(lr){10-13}
DINOv2 & SigLIP & Early L & Last L & T1: Spoon & T2: Carrot & T3: Stack & T4: Eggplant & Avg. ID & Visual & Novel Object & Novel Category & Avg. OOD \\

\midrule 
\multicolumn{4}{c|}{MiniVLA-OFT$^\dagger$} & 22.0 & 30.0 & 0.0 & 2.0 & 13.5 & 12.7 & 2.0 & 12.0 & 8.9 \\

\midrule
\textcolor{cyan}{\faSnowflake} & \textcolor{cyan}{\faSnowflake} & \textcolor{red}{\faFire} & \textcolor{red}{\faFire} & 22.0 & 24.0 & 0.0 & 34.0 & 20.0 & 20.0 & 13.3 & 10.7 & 15.0 \\  

\textcolor{red}{\faFire} & \textcolor{red}{\faFire} & \textcolor{cyan}{\faSnowflake} & \textcolor{cyan}{\faSnowflake} & 0.0 & 0.0 & 0.0 & 0.0 & 0.0 & 1.3 & 0.0 & 2.7 & 1.2 \\  

\textcolor{cyan}{\faSnowflake} & \textcolor{cyan}{\faSnowflake} & \textcolor{cyan}{\faSnowflake} & \textcolor{cyan}{\faSnowflake} & 0.0 & 0.0 & 0.0 & 0.0 & 0.0 & 0.0 & 0.0 & 0.0 & 0.0\\  

\textcolor{cyan}{\faSnowflake} & \textcolor{red}{\faFire} & \textcolor{red}{\faFire} & \textcolor{red}{\faFire} & 42.0 & 32.0 & 6.0 & 42.0 & 30.5 & 32.7 & 51.3 & 11.3 & 33.6 \\  

\textcolor{red}{\faFire} & \textcolor{cyan}{\faSnowflake} & \textcolor{red}{\faFire} & \textcolor{red}{\faFire} & 12.0 & 10.0 & 12.0 & 90.0 & 31.0 & 29.3 & 35.3 & 22.7 & 29.7 \\  

\textcolor{red}{\faFire} & \textcolor{red}{\faFire} & \textcolor{cyan}{\faSnowflake} & \textcolor{red}{\faFire} & 0.0 & 0.0 & 0.0 & 0.0 & 0.0 & 4.0 & 0.7 & 2.0 & 2.2\\  

\midrule
\multicolumn{4}{c|}{RFT} & 4.0 & 8.0 & 0.0 & 94.0 & 26.5 & 6.7 & 20.0 & 14.0 & 13.6\\

\bottomrule
\end{tabular}%
}%
\label{tab:prelim_simpler}
\vspace{-1em}
\end{table*}
\begin{table}[!h]
\centering
\caption{Comparison of freezing different parts of VLA on LIBERO. Early L and Last L denote early and the last language layers. \textcolor{cyan}{\faSnowflake} and \textcolor{red}{\faFire} represent freeze and full fine-tune.}
\resizebox{\linewidth}{!}{
\begin{tabular}{cccc|c|ccccc}
\toprule
\multicolumn{4}{c|}{\textbf{Modules}} & \textbf{ID} & \multicolumn{5}{c}{\textbf{OOD}}  \\ 
\cmidrule(lr){1-4} \cmidrule(lr){5-5} \cmidrule(lr){6-10}
DINOv2 & SigLIP & Early L & Last L & LIBERO-90 & -Spatial & -Object & -Goal & -Long & \textbf{Avg. OOD} \\ 
\midrule

\multicolumn{4}{c|}{VLA-Adapter$^\dagger$} & 83.0 & 2.0 & 5.0 & 1.0 & 2.0 & 2.5 \\

\midrule
\textcolor{cyan}{\faSnowflake} & \textcolor{cyan}{\faSnowflake} & \textcolor{red}{\faFire} & \textcolor{red}{\faFire} & 92.0 & 0.0 & 3.0 & 0.0 & 6.0 & 2.3 \\

\textcolor{red}{\faFire} & \textcolor{red}{\faFire} & \textcolor{cyan}{\faSnowflake} & \textcolor{cyan}{\faSnowflake} & 29.0 & 1.0 & 4.0 & 1.0 & 0.0 & 1.5 \\

\textcolor{cyan}{\faSnowflake} & \textcolor{cyan}{\faSnowflake} & \textcolor{cyan}{\faSnowflake} & \textcolor{cyan}{\faSnowflake} & 19.0 & 0.0 & 0.0 & 1.0 & 0.0 & 0.3 \\

\textcolor{cyan}{\faSnowflake} & \textcolor{red}{\faFire} & \textcolor{red}{\faFire} & \textcolor{red}{\faFire} & 89.0 & 0.0 & 6.0 & 0.0 & 6.0 & 3.0\\  

\textcolor{red}{\faFire} & \textcolor{cyan}{\faSnowflake} & \textcolor{red}{\faFire} & \textcolor{red}{\faFire} & 73.0 & 1.0 & 0.0 & 0.0 & 3.0 & 1.0 \\  

\textcolor{red}{\faFire} & \textcolor{red}{\faFire} & \textcolor{cyan}{\faSnowflake} & \textcolor{red}{\faFire} & 91.0 & 0.0 & 4.0 & 6.0 & 8.0 & 4.5\\

\midrule
\multicolumn{4}{c|}{RFT} & 88.0 & 1.0 & 2.0 & 0.0 & 6.0 & 2.3 \\

\bottomrule
\end{tabular}}

\label{tab:prelim_freeze}
\end{table}

\subsection{Catastrophic Forgetting in VLAs}
VLA models are well known to struggle with zero-shot generalization across new tasks, environments, and instructions. Prior work traces this largely to catastrophic forgetting during action finetuning. ReVLA \cite{dey2024revlarevertingvisualdomain} attributes the issue to collapse in the DINOv2 \cite{oquab2024dinov2learningrobustvisual} encoder, where depth estimation degrades into low-detail, spatially uniform maps -- losing the rich geometry preserved in the pretrained DINOv2 features of PrismaticVLM \cite{karamcheti2024prismaticvlmsinvestigatingdesign}. \cite{kachaev2025dontblindvlaaligning} similarly find that fine-tuned VLAs attend to irrelevant regions under OOD conditions, with t-SNE analyses revealing compressed, degenerate feature spaces. \cite{grover2025enhancinggeneralizationvisionlanguageactionmodels} further observe over-sensitivity to background variation and instruction phrasing, suggesting overfitting to language form rather than meaning. Collectively, these studies identify action finetuning as the primary source of generalization failure, motivating efforts to preserve VLM representations during adaptation.

Freezing parts of a pretrained VLM is a simple yet effective way to preserve learned representations, preventing weights from drifting in suboptimal directions during finetuning. Recent works have refined this idea in various ways: \cite{grover2025enhancinggeneralizationvisionlanguageactionmodels} freeze the vision encoder and train a separate copy, \cite{kachaev2025dontblindvlaaligning} regularize against a frozen teacher, and ReVLA \cite{dey2024revlarevertingvisualdomain} interpolates vision-encoder weights between pretrained and adapted models while tuning the language backbone. While these strategies improve generalization, they often incur extra computation and supervision. Motivated by this line of work, we investigate how far simple freezing alone can bridge the generalization gap between VLMs and VLAs.

\subsection{How good is model freezing?}
\label{sec:freeze_prelim}
To study the impact of model freezing, we decompose the VLA architecture into four components -- DINOv2, SigLIP, early language layers, and late language layers -- and systematically evaluate different freezing configurations on both ID and OOD settings. We focus on the OpenVLA \cite{kim2024openvlaopensourcevisionlanguageactionmodel} family, which integrates a joint DINOv2--SigLIP visual encoder. This modular design allows precise control over which capabilities are retained during adaptation: DINOv2 contributes depth perception and geometric priors, while SigLIP provides strong vision--language alignment. To ensure coverage across model scales and benchmarks, we conduct experiments with VLA-Adapter \cite{wang2025vlaadaptereffectiveparadigmtinyscale} on LIBERO and MiniVLA-OFT \cite{belkhale2024minivla} on SimplerEnv BridgeV2 (see Sec.~\ref{sec:main_exp} for experimental details). 

Our results (Tables~\ref{tab:prelim_simpler} and~\ref{tab:prelim_freeze}) reveal several key insights into the efficacy of freezing:  
\begin{enumerate}
    \item \textbf{Language adaptation is essential.} Freezing the language backbone severely limits adaptation to the action space, yielding near-zero performance on SimplerEnv and up to a 60\% drop on LIBERO.  
    \item \textbf{Freezing the vision encoder generally improves performance.} On SimplerEnv, ID accuracy increases by 7--17\% and OOD by 7--25\%; on LIBERO, gains are smaller but consistent (around 5\% ID). In \cref{fig:prelim_viz}, freezing vision components allows the robot arm to grasp the bowl, whereas all other configurations struggle.
    \item \textbf{Late language layers drive task performance.} Finetuning late language layers while freezing early ones boosts OOD generalization by 1--3\%, aligning with their direct role in producing action outputs.  
    \item \textbf{Preserving DINOv2 matters more than SigLIP.} Freezing DINOv2 yields roughly 5\% higher OOD performance than freezing SigLIP, underscoring the importance of maintaining strong geometric priors.  
    \item \textbf{Freezing effects are not universally consistent.} Freezing demonstrates improvements over full fine-tuning. Yet, some configurations (ex. freeze early language) improve one benchmark but harm another, indicating that freezing introduces task-dependent inductive biases.  
\end{enumerate}

Overall, our findings suggest a hierarchy of importance among modules:
\[
\boxed{\text{DINOv2} > \text{SigLIP} > \text{Early language} > \text{Late language}}
\]

Yet, \textbf{freezing alone is not a reliable path to generalization}. While it allows selective preservation of pretrained knowledge, it also imposes strong biases that can degrade performance in unseen settings. A more robust alternative lies in soft regularization, where finetuning is guided rather than restricted.

\subsection{From hard constraints to soft constraints}

The robust finetuning (RFT) literature (see \cref{sec:spd}) offers a natural foundation for our goal of constraining weight updates while preserving pretrained representations. In principle, RFT lowers task loss while keeping parameter drift minimal. While the intuition fits seamlessly with our goals, we find that naively applying RFT directly on the action finetuning offers only modest improvements over the baseline in both SimplerEnv (\cref{tab:prelim_simpler}) and LIBERO (\cref{tab:prelim_freeze}). This is because RFT techniques impose a single hyperparameter constraint across the entire model, meaning that layers either constrict closer towards or expand away from 
the pretrained weights at a uniform rate. As shown at the bottom of \cref{fig:vis_l2_dist}, this results in all components of the VLM diverging equal amounts from their pretrained representations. As we observed from \cref{sec:freeze_prelim}, however, different components should evolve at different rates -- some should be allowed more flexibility to adapt to the action space, whereas others should be constrained to preserve their pretrained priors. 

\subsection{Module-Wise Proximity Scheduling}
\label{sec:maps}
Our preliminary studies highlights a key tradeoff: model freezing offers high modularity at the cost of heavy inductive bias, RFT offers lighter inductive bias but has no modularity. To bridge the two paradigms and successfully adapt robust finetuning for the action domain, we introduce a new framework of RFT methods specifically tailored for VLA models, known as \textbf{Module-Wise Proximity Scheduling (MAPS).}

MAPS assigns distinct proximity schedules to different components of the base VLM, enabling strong preservation where pretrained structure is crucial (e.g., early vision and early language layers) while allowing more aggressive adaptation where grounding to actions is most beneficial (e.g., higher-level language layers and the action head). In other words, MAPS is designed to \emph{learn when to preserve and when to adapt}, avoiding the rigidity of full freezing and the instability of uniform finetuning.

Concretely, MAPS modifies SPD by replacing its \emph{global} proximity hyperparameter~$\lambda$ with a \emph{layer-wise}, architecture-aware schedule. Let $\mathcal{L} = (\ell_1,\ldots,\ell_{|\mathcal{L}|})$ denote the ordered stack of model submodules (\textsc{DINOv2} $\rightarrow$ \textsc{SigLIP} $\rightarrow$ \textsc{Bridge} $\rightarrow$ \textsc{Language}). For each layer index $k$, MAPS assigns a proximity weight \[\lambda_k \;=\; \lambda_{\max}\!\left(1 - \frac{k-1}{|\mathcal{L}|-1}\right),\] yielding a linear decay from $\lambda_1 = \lambda_{\max}$ for early visual layers to $\lambda_{|\mathcal{L}|} = 0$ for language layers. In addition, MAPS assigns $\lambda_k = 0$ for all modules that are initialized from scratch, such as the action head, proprioceptive-state projector, and other non-pretrained components. Since these modules have no pretrained parameters $\theta_0$ to preserve, MAPS simply performs full fine-tuning on them.

During optimization, let $\theta$ denote the parameters of the current layer, $\theta_0$ their pretrained initialization, and $g_t$ the gradient at iteration $t$. MAPS first computes an unconstrained step $\widetilde{\theta}_t$, after which it evaluates the gradient--displacement correlation \[c_t \;\coloneqq\; -\, g_t^{\!\top}\bigl(\theta_{t-1}-\theta_0\bigr).\] Whenever $c_t < 0$, indicating that the update direction is inconsistent with preserving pretrained structure, MAPS applies a projection toward~$\theta_0$ with \emph{layer-specific} strength:
\[
\theta_t 
\;\leftarrow\;
\widetilde{\theta}_t
\;-\;
\lambda_k\, r_t\!\bigl(\widetilde{\theta}_t - \theta_0\bigr),
\]
where $r_t$ is the SPD deviation ratio (see \cref{sec:spd}).  

\begin{figure}[!h]
    \centering
    \includegraphics[width=0.8\linewidth]{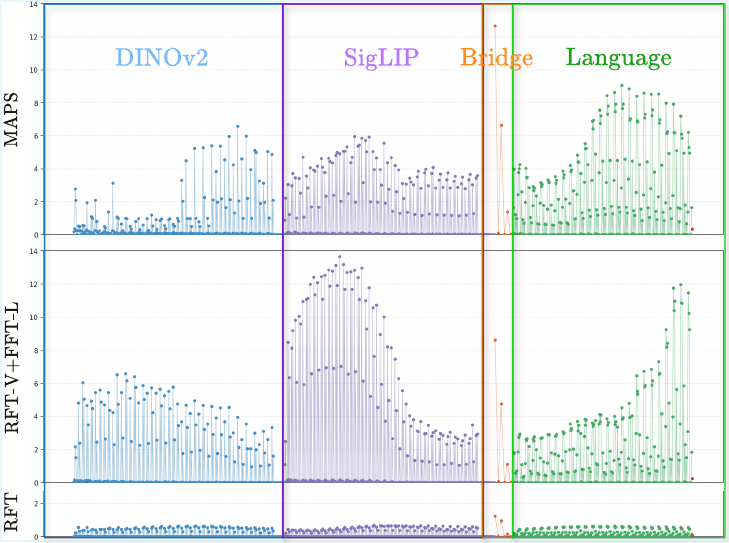}
    \caption{We calculate the $\ell_2$ distance between fine-tuned and pre-trained weights. MAPS produces a smooth, module-aware decay in deviation from pretrained initialization (top). Applying robust finetuning only to the vision stack and full finetuning on language (RFT-V+FFT-L) constrains DINOv2 heavily but leaves SigLIP and language relatively unconstrained (middle). Uniform $\lambda$ (RFT) constrains all layers equally (bottom).}
    \label{fig:vis_l2_dist}
\end{figure}

Unlike SPD, which uses a single global constraint and therefore cannot differentiate between layers with distinct gradient scales or semantic roles, MAPS explicitly encodes architectural structure: visual layers remain strongly anchored to pretrained perceptual representations, while deeper language layers are allowed to adapt more flexibly to task-level semantics. This is highlighted in \cref{fig:vis_l2_dist}, where DINOv2 weights diverge least, followed by SigLIP weights, and finally language weights adapt most. As shown in \cref{sec:ablation}, the linear schedule outperforms constant and cosine alternatives, underscoring the importance of structured, layer-wise proximity control.

MAPS is lightweight and plug-and-play: it introduces no additional parameters, requires no auxiliary networks or extra data, and can be applied to existing VLA architectures without modification. By directly encoding the empirical hierarchy revealed in our preliminary studies, MAPS strikes an effective balance between retaining essential pretrained priors and adapting to new action domains.
\section{Experiments}
\label{sec:main_exp}

We evaluate MAPS through three guiding questions. \textbf{(RQ1)} How well does it generalize across ID and OOD tasks? To test this, we use three diverse benchmarks -- SimplerEnv \cite{li2024evaluatingrealworldrobotmanipulation} (\cref{sec:main_exp_simpler}), CALVIN \cite{mees2022calvinbenchmarklanguageconditionedpolicy} (\cref{sec:main_exp_calvin}), and LIBERO \cite{liu2023liberobenchmarkingknowledgetransfer} (\cref{sec:main_exp_libero}) -- that span semantic, environmental, and visual shifts, providing a comprehensive measure of robustness. \textbf{(RQ2)} Are MAPS’ gains consistent across architectures? We evaluate four state-of-the-art VLA backbones -- MiniVLA-VQ, MiniVLA-OFT, VLA-Adapter, and OpenVLA-OFT -- to test architecture-agnostic benefits.  \textbf{(RQ3)} Does the robustness afforded by MAPS transfer to real-world deployment? We assess real-world transfer by deploying MAPS on a Franka Emika Panda manipulator (\cref{sec:main_exp_franka}). Importantly, across all benchmarks, \textit{we evaluate MAPS by fine-tuning the base VLM weights directly}. In other words, we initialize each VLA backbone from its pretrained VLM and train it from scratch using MAPS.

\noindent \textbf{Backbones.} Because MAPS does not rely on architecture-specific hyperparameters or task-specific loss modifications, it can be dropped into models with any visual encoder, language model, or action tokenization strategy. To demonstrate this flexibility, we instantiate MAPS across four widely used VLA backbones: MiniVLA-VQ \cite{belkhale2024minivla}, MiniVLA-OFT, OpenVLA-OFT \cite{kim2025finetuningvisionlanguageactionmodelsoptimizing}, and VLA-Adapter \cite{wang2025vlaadaptereffectiveparadigmtinyscale}. MiniVLA pairs a DINOv2-SigLIP visual stack with a Qwen2.5-0.5B \cite{bai2025qwen25vltechnicalreport} language model; the VQ variant replaces the typical uniform discretized action bins with a vector-quantized VAE \cite{lee2024behaviorgenerationlatentactions}, enabling multi-step action chunking from a single forward pass. OpenVLA-OFT instead uses a LLaMA-2 7B \cite{touvron2023llama2openfoundation} backbone and a parallel decoding head that directly regresses continuous actions via an $\ell_1$ loss, fully avoiding discretization. By implementing this OFT-style action head on the MiniVLA backbone, we also create a lightweight MiniVLA-OFT variant, allowing us to test our method with both small and large LMs while holding other components fixed. Finally, VLA-Adapter introduces a lightweight action head with Bridge Attention, enabling efficient and low-cost VLA training.

\subsection{Performance on SimplerEnv}
\label{sec:main_exp_simpler}
\begin{table*}[!ht]
\centering
\caption{\textbf{SimplerEnv Results}. ID includes 4 tasks following SimplerEnv-Bridge setup. OOD evaluation suites include Visual, Novel Object, and Novel Category. $^*$ model checkpoints borrowed from original papers. $^\dagger$ models pretrained from scratch on SimplerEnv.}
\vspace{-1em}
\resizebox{1.0\linewidth}{!}{%
\begin{tabular}{lccccccccc}
\toprule
\multirow{2}{*}{\textbf{Model}} & \multicolumn{5}{c}{\textbf{ID}} & \multicolumn{4}{c}{\textbf{OOD}} \\
\cmidrule(lr){2-6} \cmidrule(lr){7-10}
& T1: Spoon & T2: Carrot & T3: Stack & T4: Eggplant & Avg. ID & Visual & Novel Object & Novel Category & Avg. OOD \\

\midrule
RT-1-X \cite{brohan2023rt1roboticstransformerrealworld} & 4.2 & 0.0 & 0.0 & 0.0 & 1.1 & 0.0 & 4.0 & 6.1 & 3.4 \\
Octo \cite{octomodelteam2024octoopensourcegeneralistrobot, fan2025interleavevlaenhancingrobotmanipulation} & 12.5 & 41.7 & 15.8 & 0.0 & 17.5 & 12.6 & 10.8 & 8.4 & 10.6 \\
$\pi_0$ \cite{black2024pi0visionlanguageactionflowmodel} & 52.5 & 87.9 & 83.8 & 52.5 & 69.2 & 71.4 & 30.2 & 21.0 & 40.9 \\

\midrule
MiniVLA-VQ $^\dagger$ \cite{belkhale2024minivla} & 52.0 & 24.0 & 42.0 & 58.0 & 44.0 & 37.3 & 34.0 & 27.3 & 32.9 \\  
\rowcolor{gray!15} \textbf{+MAPS} (\textit{Ours}) & 44.0 & 30.0 & 76.0 & 20.0 & 42.5 & 48.0 & 26.0 & 29.3 & 34.4 \\
\rowcolor{gray!15} \quad \quad  \quad $\Delta$ & 
\textcolor{red}{-8.0} & \textcolor{green!70!black}{+6.0} & 
\textcolor{green!70!black}{+34.0} & \textcolor{red}{-38.0} & \textcolor{red}{-1.5} & \textcolor{green!70!black}{+10.7} & \textcolor{red}{-8.0} & \textcolor{green!70!black}{+2.0} & \textcolor{green!70!black}{+1.5}  \\ 

\midrule 
MiniVLA-OFT$^\dagger$ & 22.0 & 30.0 & 0.0 & 2.0 & 13.5 & 12.7 & 2.0 & 12.0 & 8.9\\
\rowcolor{gray!15} \textbf{+MAPS} (\textit{Ours}) & 34.0 & 38.0 & 0.0 & 48.0 & 30.0 & 32.0 & 48.7 & 26.7 & 35.8\\
\rowcolor{gray!15} \quad \quad  \quad $\Delta$ & \textcolor{green!70!black}{+12.0} & \textcolor{green!70!black}{+8.0} & +0.0 & \textcolor{green!70!black}{+46.0} & \textcolor{green!70!black}{+16.5} & \textcolor{green!70!black}{+19.3} & \textcolor{green!70!black}{+46.7} & \textcolor{green!70!black}{+14.7} & \textcolor{green!70!black}{+26.9} \\ 

\midrule 
OpenVLA-OFT$^\dagger$ \cite{kim2025finetuningvisionlanguageactionmodelsoptimizing} & 16.0 & 18.0 & 0.0 & 58.0 & 23.0 & 16.0 & 9.3 & 0.7 & 8.7 \\ 
\rowcolor{gray!15} \textbf{+MAPS} (\textit{Ours}) & 2.0 & 16.0 & 0.0 & 74.0 & 23.0 & 18.7 & 20.0 & 12.7 & 17.1 \\  
\rowcolor{gray!15} \quad \quad  \quad $\Delta$ & \textcolor{red}{-14.0} & \textcolor{red}{-2.0} & +0.0 & \textcolor{green!70!black}{+16.0} & +0.0 & \textcolor{green!70!black}{+2.7} & \textcolor{green!70!black}{+10.7} & \textcolor{green!70!black}{+12.0} & \textcolor{green!70!black}{+8.4} \\ 
\bottomrule
\end{tabular}%
}%
\label{tab:simpler_result}
\vspace{-1em}
\end{table*}

\begin{figure*}[!h]
    \centering
    \includegraphics[width=\linewidth]{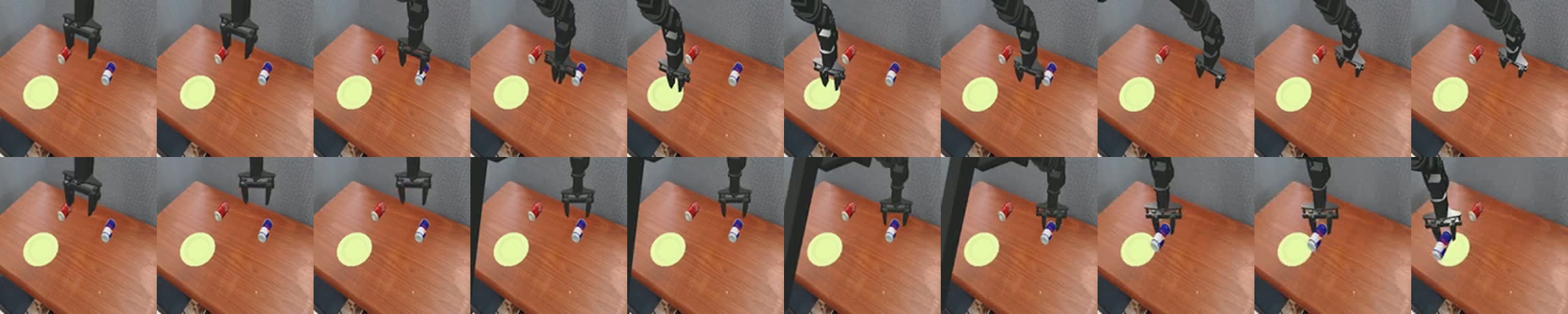}
    \caption{SimplerEnv qualitative example (new object: Red Bull can). Top: Full fine-tuning fails to complete the task. Bottom: MAPS successfully places the can on the plate.}
    \label{fig:simplerenv_new_object_qualitative}
    \vspace{-5mm}
\end{figure*}

\noindent \textbf{Experimental setting.} SimplerEnv is a lightweight simulation benchmark for efficient robotic manipulation evaluation. Because its tasks are originally text-only, we follow \cite{fan2025interleavevlaenhancingrobotmanipulation} to add visual observations and align it with the VLA setting. For ID evaluation, we use the standard WidowX BridgeData V2 Visual Matching setup and report performance on four tasks: spoon on towel, carrot on plate, stack cube, and put eggplant in basket. For OOD evaluation, we adopt the extended suite from \cite{fan2025interleavevlaenhancingrobotmanipulation}, which includes: (1) Visual Generalization, placing three ID tasks in unseen environments; (2) Novel Object, introducing new instances from known categories; and (3) Novel Category, involving entirely new object types. All VLA backbones are pretrained from scratch on the ID tasks, and we report success rates on both ID and OOD splits (Table \ref{tab:simpler_result}). We also include prior VLA baselines, though these benefit from large-scale pretraining before fine-tuning on SimplerEnv.

\noindent \textbf{Evaluation results.} On MiniVLA-OFT, MAPS matches the ID performance of Vanilla finetuning, but demonstrates 10-20\% improvement on OOD generalization. Notably, despite a substantially smaller pretraining corpora of just the BridgeData V2 Visual Matching tasks, MAPS matches and even surpasses the OOD generalizability of much larger and much more extensively pretrained baselines such as RT-1-X, Octo and $\pi_0$. On MiniVLA-VQ, MAPS maintains similar ID performance and offers a moderate increase in OOD performance. On OpenVLA-OFT, MAPS maintains similar ID performance and improves average OOD by 8.4\%. Again, this suggests that MAPS scales effectively across a variety of backbone sizes and architectures (RQ2). 

\subsection{Performance on CALVIN}
\label{sec:main_exp_calvin}
\begin{figure}[t]
    \centering
    \includegraphics[width=0.8\linewidth]{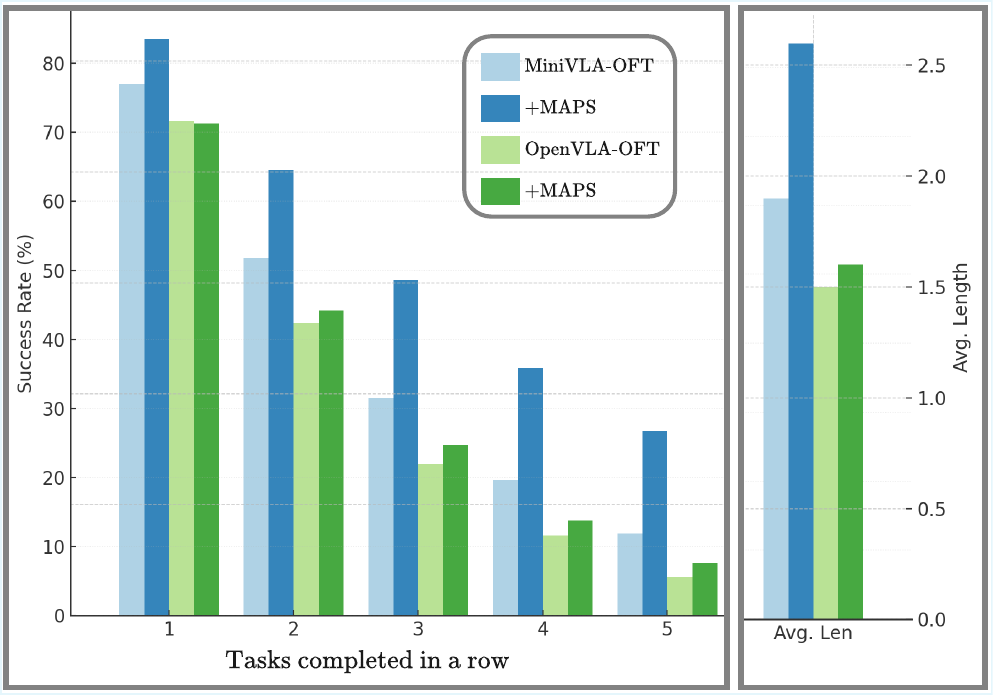}
    \caption{\textbf{CALVIN Results}. MAPS improves success rates across all action horizons (25\% for MiniVLA-OFT and 3\% for OpenVLA-OFT) and average length (+0.7 for MiniVLA-OFT and +0.1 for OpenVLA-OFT).}
    \label{fig:calvin_results}
    \vspace{-1em}
\end{figure}

\noindent \textbf{Experimental setting.} CALVIN features long-horizon tabletop manipulation tasks in a real-to-sim setting, partitioned into the A, B, C, and D splits that increasingly withhold task--environment combinations. We follow the standard ABC$\rightarrow$D protocol: models are trained on splits A--C and evaluated OOD on split D. Evaluation uses CALVIN’s long-horizon, multi-task language control setup, where agents must complete sequences of five instructions; each new instruction is provided only after the previous one is completed, with success defined by reaching the instruction’s goal state.
The benchmark reports two metrics: \emph{tasks completed in a row}, the success rate for completing exactly $k$ consecutive tasks ($k=1$--$5$), and \emph{average length (avg.\ len)}, the mean number of tasks completed per five-instruction sequence across all episodes. 

\noindent \textbf{Evaluation results.} As shown in \cref{fig:calvin_results}, MAPS leads to a +0.7 gain in average sequence length and consistently improves success rates across all task horizons on MiniVLA-OFT. The largest gains are observed for longer horizons: MAPS improves the success rate of completing 2-, 3-, 4-, and 5-task sequences by roughly 15\% each, while the 1-task success rate increases by 7\%. MAPS demonstrates modest improvements over OpenVLA-OFT, boosting success rates by 2\% across task horizons. Notably, MAPS demonstrates larger gains on stronger baselines, suggesting that it complements rather than replaces existing capabilities, and scales effectively with model quality (RQ2).

\subsection{Performance on LIBERO}
\label{sec:main_exp_libero}
\begin{table}[!h]
\centering
\caption{\textbf{LIBERO Results}. $^*$ models borrowed from \cite{belkhale2024minivla}. $^\dagger$ models pretrained from scratch on LIBERO-90.  $^\ddagger$ models finetuned on LIBERO-90.}
\resizebox{\linewidth}{!}{
\begin{tabular}{lcccccc}
\toprule
\multirow[c]{2}{*}{\textbf{Model}}& \multicolumn{1}{c}{\textbf{ID (\%)}} & \multicolumn{5}{c}{\textbf{OOD (\%)}}  \\ 
\cmidrule(lr){2-2} \cmidrule(lr){3-7}
  & LIBERO-90 & -Spatial & -Object & -Goal & -Long & \textbf{Avg. OOD} \\ 
\midrule
OpenVLA &61.4 &- &- &- &- &- \\ 
miniVLA$^*$  & 62.0 &0.0 &0.0 &0.0 &1.0 & 0.25 \\ 
miniVLA-VQ$^*$ & 77.0 &0.0 &0.0  &3.0 &1.0 & 1.0\\ 
miniVLA-hist.$^*$ & 82.0 & 0.0&1.0 &2.0 &7.0 & 2.5\\
miniVLA-wrist$^*$ & 82.1& 0.0& 0.0&0.0 &0.0 &0.0 \\
SpatialVLA$^\ddagger$ & 46.2 & 0.0 & 0.0 & 0.67 &1.33 &0.5 \\

\midrule 
MiniVLA-VQ$^\dagger$ & 79.0 & 0.0 & 0.0 & 0.0 & 0.0 & 0.0 \\ 
\rowcolor{gray!15} \textbf{+MAPS} (\textit{Ours}) & 82.0 & 0.0 & 1.0 & 8.0 &10.0 & 4.75 \\  
\rowcolor{gray!15} \quad \quad  \quad $\Delta$  & \textcolor{green!70!black}{+{3.0}} &\textcolor{black}{+{0.0}} &\textcolor{green!70!black}{+{1.0}}  &\textcolor{green!70!black}{+{8.0}} &\textcolor{green!70!black}{+{10.0}} &\textcolor{green!70!black}{+{4.75}} \\ 

\midrule 
MiniVLA-OFT$^\dagger$ & 91.0 & 0.0 & 8.0 & 7.0 & 4.0 & 4.75 \\ 
\rowcolor{gray!15} \textbf{+MAPS} (\textit{Ours}) & 91.0 & 6.0 & 14.0 & 0.0 &8.0 & 7.0 \\  
\rowcolor{gray!15} \quad \quad  \quad $\Delta$  & \textcolor{black}{+{0.0}} &\textcolor{green!70!black}{+{6.0}} &\textcolor{green!70!black}{+{6.0}}  &\textcolor{red}{-{7.0}} &\textcolor{green!70!black}{+{4.0}} &\textcolor{green!70!black}{+{2.25}} \\ 

\midrule 
VLA-Adapter$^\dagger$ & 83.0 & 2.0 & 5.0 & 1.0 & 2.0 & 2.5 \\ 
\rowcolor{gray!15} \textbf{+MAPS} (\textit{Ours}) & 88.0 & 0.0 & 7.0 & 6.0 &6.0 & 4.75 \\  
\rowcolor{gray!15} \quad \quad  \quad $\Delta$  & \textcolor{green!70!black}{+{5.0}} &\textcolor{red}{-{2.0}} &\textcolor{green!70!black}{+{2.0}}  &\textcolor{green!70!black}{+{5.0}} &\textcolor{green!70!black}{+{4.0}} &\textcolor{green!70!black}{+{2.25}} \\ 

\midrule 
OpenVLA-OFT$^\dagger$ & 92.0 & 3.0 & 9.0 & 3.0 & 5.0 & 5.0 \\ 
\rowcolor{gray!15} \textbf{+MAPS} (\textit{Ours}) & 90.0 & 0.0 & 11.0 & 1.0 & 10.0 & 5.5 \\  
\rowcolor{gray!15} \quad \quad  \quad $\Delta$ & \textcolor{red}{-{2.0}} & \textcolor{red}{-{3.0}} & \textcolor{green!70!black}{+{2.0}} &\textcolor{red}{-{2.0}} & \textcolor{green!70!black}{+5.0} & \textcolor{green!70!black}{+0.5} \\ 

\bottomrule
\end{tabular}}
\vspace{-1em}
\label{tab:libero_result}
\end{table}

\noindent \textbf{Experimental setting.} 
The LIBERO benchmark consists of procedurally generated household manipulation tasks designed to evaluate generalization across task configurations, object appearances, and spatial layouts. It includes several predefined splits -- LIBERO-90, LIBERO-Long, LIBERO-Goal, LIBERO-Spatial, and LIBERO-Object -- each targeting a specific type of generalization, such as task diversity, long-horizon planning, unseen goals, spatial rearrangements, or novel objects. Agents are evaluated in a reset-free setting and must complete multi-step tasks from natural language instructions.

Previous works often rely on large-scale external pretraining -- such as on Open X Embodiment -- followed by fine-tuning separate model variants for each LIBERO split, and reporting results using these tailored models. In contrast, \textbf{our approach uses no external pretraining whatsoever}. Instead, we follow the more challenging protocol of \cite{wang2025vlaadaptereffectiveparadigmtinyscale}: we pre-train solely on the LIBERO-90 subset, and evaluate both ID performance on LIBERO-90 and OOD generalization on LIBERO-Spatial, LIBERO-Object, LIBERO-Goal, and LIBERO-Long. We report the success rates using 10 trials per task in Table \ref{tab:libero_result}.

\noindent \textbf{Evaluation results.} MAPS delivers consistent improvements in OOD performance across all four model backbones, yielding an average gain of 1–5\%. Notably, the base MiniVLA model shows no inherent OOD generalization capability; however, with MAPS, its performance increases by 8\% on LIBERO-Goal and 10\% on LIBERO-Long. For MiniVLA-OFT, MAPS enhances generalization on LIBERO-Spatial and LIBERO-Object by 6\%. Similarly, MAPS improves the VLA-Adapter’s performance on LIBERO-Goal and LIBERO-Long by 5\% and OpenVLA-OFT's performance on LIBERO-Long by 5\%. Beyond OOD gains, MAPS also provides moderate improvements on ID tasks, boosting MiniVLA-VQ performance by 3\% and VLA-Adapter by 5\%. This suggests that MAPS' gains are indeed consistent across architectures and generalize well to different action tokenization schemes (RQ2). These results, in conjunction with results on the other benchmarks, also suggest that MAPS generalizes effectively across diverse ID and OOD tasks (RQ1).

\subsection{Performance on Franka Emika Panda}
\label{sec:main_exp_franka}
\begin{table*}[!h]
\centering
\caption{\textbf{Franka Results}.}
\vspace{-1em}
\resizebox{1.0\linewidth}{!}{%
\begin{tabular}{lcccccccccccc}
\toprule
\multirow{2}{*}{\textbf{Model}} & \multicolumn{5}{c}{\textbf{ID}} & \multicolumn{5}{c}{\textbf{OOD}} \\
\cmidrule(lr){2-6} \cmidrule(lr){7-11}
& T1: Coke & T2: Block & T3: Cup & T4: Laptop & Avg. ID & T1: Coke & T2: Block & T3: Cup & T4: Laptop & Avg. OOD \\

\midrule 
MiniVLA-OFT$^\dagger$ & 30.0 & 30.0 & 0.0 & 100.0 & 40.0 & 50.0 & 20.0 & 0.0 & 30.0 & 22.5 \\
\rowcolor{gray!15} \textbf{+MAPS} (\textit{Ours}) & 50.0 & 70.0 & 80.0 & 90.0 & 72.5 & 100.0 & 30.0 & 20.0 & 60.0 & 52.5\\
\rowcolor{gray!15} \quad \quad  \quad $\Delta$ & \textcolor{green!70!black}{+20.0} & \textcolor{green!70!black}{+40.0} & \textcolor{green!70!black}{+80.0} & \textcolor{red}{-10.0} & \textcolor{green!70!black}{+32.5} & \textcolor{green!70!black}{+50.0} & \textcolor{green!70!black}{+10.0} & \textcolor{green!70!black}{+20.0} & \textcolor{green!70!black}{+30.0} & \textcolor{green!70!black}{+30.0} \\ 

\bottomrule
\end{tabular}%
}%
\label{tab:franka_result_main}
\vspace{-1em}
\end{table*}

\noindent \textbf{Experimental setting.} We conduct real-world tabletop manipulation experiments using a single Franka Emika Panda robotic arm. We form a pretraining corpora of 600 demonstrations across 4 different ID training tasks following \cite{Coholich2025Sim2Real}, which are:
\begin{enumerate}
    \item Coke ID: grab the coke can and lift it up,
    \item Blocks ID: stack the blue block on the green block,
    \item Cups ID: stack the orange cup onto the green cup,
    \item Laptop ID: close the laptop lid fully.
\end{enumerate}
We train MiniVLA-OFT (directly updating the VLM weights) using our self-collected demonstration dataset with both vanilla finetuning and MAPS.

To assess generalization, we evaluate the trained policies on the 4 ID tasks as well as on 4 corresponding OOD variants. The OOD tasks introduce controlled distribution shifts while preserving the underlying manipulation objectives. Specifically:
\begin{enumerate}
    \item Coke OOD: coke can is replaced with red bull can;
    \item Blocks OOD: blue block is elevated by placing an additional block underneath it;
    \item Cups OOD: orange and green cups are replaced with yellow and blue cups, respectively;
    \item Laptop OOD: Alienware laptop is replaced with a MacBook.
\end{enumerate}

Overall, we design the OOD tasks to systematically probe performance under diverse distribution shifts. Coke OOD introduces novel instruction, class (Red Bull can), object geometry, and color. Blocks OOD tests novel elevation (elevated blue block). Cups OOD tests novel instruction and color. Laptop OOD tests novel class, color, object geometry, and altered physics (joint stiffness). For each task, including both ID and OOD variants, we conduct 10 evaluation trials using two object-location configurations, with 5 trials per configuration. All evaluations are executed in the real-world setup using the same Franka system as in training with a max limit of 500 timesteps. We present our quantitative results below. We also provide qualitative results, rollout visualizations, setup details, and an in-depth results discussion along with observed success and failure modes in Appendix \cref{sec:suppl_franka}.

\noindent \textbf{Evaluation results.} \Cref{tab:franka_result_main} presents the success rates for each task together with the overall average performance. MAPS yields substantial gains over MiniVLA-OFT on both ID and OOD tasks. On the ID tasks, MAPS raises average performance from 40.0\% to 72.5\%, demonstrating much stronger policy learning from limited demonstrations. MAPS also improves generalization under distribution shift, boosting average OOD success from 22.5\% to 52.5\%. Overall, MAPS consistently enhances robustness and task success across diverse manipulation scenarios, showing clear benefits for real-world policy training.

\begin{figure*}[!h]
    \centering
    \includegraphics[width=0.7\linewidth]{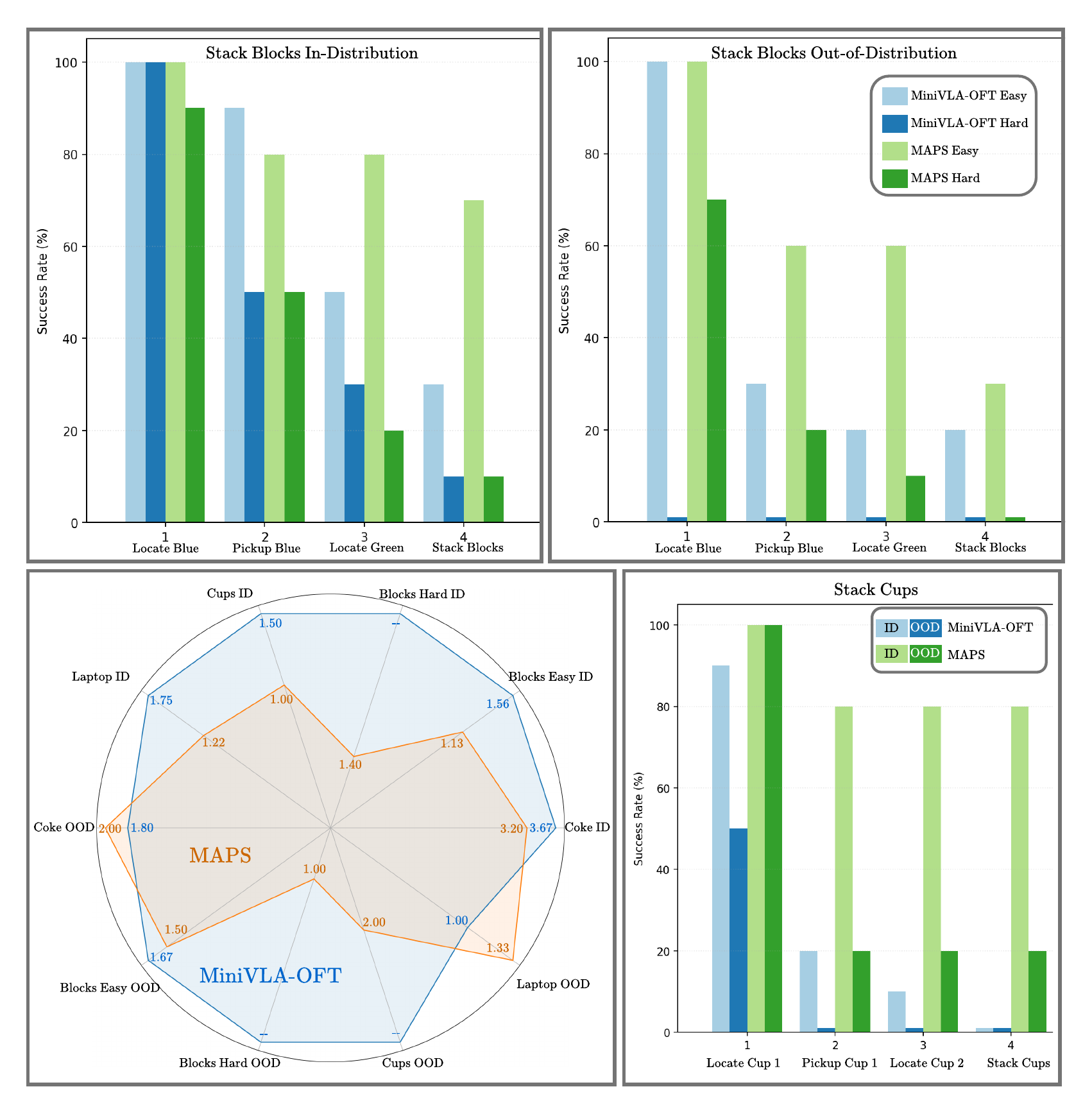}
    \caption{\textbf{Franka Detailed Results. Top Left:} Success rates for the Blocks ID Easy/Hard tasks. \textbf{Top Right:} Success rates for the Blocks OOD Easy/Hard tasks. \textbf{Bottom Left:} Average number of tries per success across all Franka tasks. \textbf{Bottom Right:} Success rates for the Cups ID/OOD tasks.}
    \vspace{-1.5em}
    \label{fig:franka_fine}
\end{figure*}

\noindent \textbf{Fine-grained task analysis.}
Beyond reporting aggregate success rates, we also provide a more fine-grained analysis of MAPS in \cref{fig:franka_fine}, differentiating non-trivial competency on intermediate steps. To this end, we decompose the multi-step Task 2: Block (\cref{fig:franka_fine} Top) and Task 3: Cup (\cref{fig:franka_fine} Bottom Right) into a sequence of subtasks: (1) locating object 1, (2) picking up object 1, (3) locating object 2, and (4) stacking object 1 on object 2. We record the success rates on each of these subtasks for a closer look at potential failure modes. For Task 2 specifically, we additionally introduce a "Hard" setting (see Appendix \cref{sec:suppl_franka_experiment}) in which the block configuration is made more challenging than the "Easy" setting reported in Table 6. In the Hard ID variant, the distance between the two blocks is increased, while in the Hard OOD variant, object 1 (the blue block) is replaced with a red block. Finally, we also track how many attempts each method needs to complete a rollout (\cref{fig:franka_fine} Bottom Left). For Task 1 and Task 4, which have no intermediate subtasks, this value is simply the number of tries until the task succeeds. For Task 2 and Task 3, which include multiple subtasks, we record attempts only for the grasping subtask, as it is the only stage where the policy can retry. 

MAPS’s fine-grained evaluations show that its biggest advantages come from more reliable grasps, more accurate intermediate localization, and more stable multi-step execution across all object categories. In Blocks ID (Easy), MAPS strengthens later-stage performance, improving second-block localization (80\% vs. 50\%), more than doubling final placements (70\% vs. 30\%), and reducing grasp attempts (1.13 vs. 1.56). Similar gains appear in Blocks OOD (Easy), boosting first-block grasping (60\% vs. 30\%), second-block localization (60\% vs. 20\%), and final placement (30\% vs. 20\%), indicating robustness to modest distribution shift. Under Hard settings, both models degrade, but MAPS still reduces repeated grasps on Blocks ID (1.4 vs. 4.2). In Blocks OOD (Hard), where object 1 is fully OOD, the baseline fails entirely while MAPS maintains non-trivial competence -- 70\% first-block localization, 20\% initial grasps, and 10\% second-block localization -- showing resilience to combined geometric and appearance shifts. For cup stacking, MAPS substantially improves every sub-stage: grasp-first success (80\% vs. 20\%), second-object localization (80\% vs. 10\%), final stacking (80\% vs. 0\%), and grasp attempts (1.0 vs. 1.5) in Cups ID; and in Cups OOD, the baseline fails beyond initial localization while MAPS successfully grasps, relocates, and stacks (2 trials each), demonstrating clear generalization. For laptop closing, both models locate and push the lid, but MAPS excels on the harder metric of fully closing: in Laptop ID it remains near-perfect (90\% vs. 100\%) with fewer attempts (1.22 vs. 1.75), and in Laptop OOD it doubles the fully closed rate (60\% vs. 30\%), showing better control accuracy under appearance change. For coke can lifting, MAPS improves success in both ID (50\% vs. 30\%) and OOD (100\% vs. 50\%), with slightly fewer attempts in ID (3.20 vs. 3.67); its perfect OOD score reflects strong appearance invariance. Overall, MAPS not only boosts final task success but consistently improves intermediate sub-task reliability across both ID and OOD conditions.
\vspace{-2mm}
\section{Ablation Studies}
\label{sec:ablation}

We design our ablation studies to rigorously evaluate our design choices and compare against prior approaches. Our ablation of different projection strength hyperparameter schedulers is shown below. For our systematic comparison against other methods for preserving pretrained representations, please see Appendix \cref{sec:suppl_ablation}.

\noindent \textbf{Comparison with different schedulers.} 
We ablate the choice of scheduler and projection strength values in Table \ref{tab:abl_scheduler_values} on the LIBERO benchmark. Specifically, we compare against regular robust-finetuning (Constant scheduler), cosine scheduler and linear scheduler on projection strength values of 0.5, 2, and 3. We find that in general, project-strength modulation offers ID and OOD benefits compared to regular robust-finetuning. We also find that the linear scheduler offers most stable ID performance and greatest gains in OOD, with the best configuration being a linear scheduler with projection strength $v=0.5$.
\vspace{-1em}

\begin{table}[!h]
\centering
\caption{\textbf{Scheduler Comparison}: Linear (MAPS, Ours) vs. Cosine (and Constant) across projection-strength values.}
\resizebox{\linewidth}{!}{
\begin{tabular}{lcccccc}
\toprule
\multirow{2}{*}{\textbf{Scheduler / Value}} & \textbf{ID} & \multicolumn{5}{c}{\textbf{OOD}}\\
\cmidrule(lr){2-7}
& LIBERO-90 & -Spatial & -Object & -Goal & -Long & \textbf{Avg. OOD} \\
\midrule
Constant ($v{=}0.5$) & 80 & 0 & 0 & 0 & 6 & 1.5 \\
Constant ($v{=}2$) & 67 & 0 & 0 & 0 & 4 & 1  \\
Constant ($v{=}3$) & 51 & 0 & 0 & 0 & 0 & 0\\
\midrule
Cosine ($v{=}0.5$)  & 81 & 0 & 2 & 0 & 6 & 2 \\
Cosine ($v{=}2$)  & 85 & 0 & 7 & 3 & 3 & 3.25 \\
Cosine ($v{=}3$) & 87 & 0 & 9 & 1 & 2 & 3 \\
\midrule
\rowcolor{gray!15} Linear ($v{=}0.5$)  & 88 & 0 & 7 & 6 & 6 & 4.75 \\
Linear ($v{=}2$) & 84 & 2 & 6 & 0 & 4 & 3 \\
Linear ($v{=}3$) & 88 & 1 & 2 & 0 & 6 & 2.25 \\
\bottomrule
\vspace{-2em}
\end{tabular}}
\label{tab:abl_scheduler_values}
\end{table}

\vspace{-0.6em}
\section{Conclusion}
\label{sec:conclusion}
\vspace{-0.5em}

We present Module-Wise Proximity Scheduling (MAPS), a robust finetuning framework designed specifically for adapting pretrained VLM backbones to action data. By uncovering a clear hierarchy in how different model components should be constrained, MAPS applies a linear schedule that selectively regularizes each module. This yields conservative updates for visual layers -- preserving essential geometric priors -- while allowing more flexible adaptation in language components. Across SOTA backbones and standard benchmarks, MAPS delivers consistent improvements, enhancing both ID and OOD performance. Our findings highlight guided proximity to pretrained VLMs as a simple yet powerful principle for scalable VLA adaptation.

\vspace{0.5em}
\noindent\textbf{Acknowledgements.} The authors gratefully acknowledge Jeremiah Coholich for his early contributions to the development of the Franka setup as well as support with the real robot experiments. This material is based upon work partially supported by the National Science Foundation under Grant No. 2239292. This research was also supported in part by Lambda, Inc. and through research cyberinfrastructure resources and services provided by the Partnership for an Advanced Computing Environment (PACE) at the Georgia Institute of Technology, Atlanta, Georgia, USA.

\clearpage
{
    \small
    \bibliographystyle{unsrtnat}
    \bibliography{main}
}

\newpage
\setcounter{page}{1}
\maketitlesupplementary

\section{Performance on Real-World Franka Emika Panda Robot}
\label{sec:suppl_franka}
\subsection{Experiment Setting}
\label{sec:suppl_franka_experiment}
\begin{figure}[!h]
    \vspace{-2em}
    \centering
    \includegraphics[width=\linewidth]{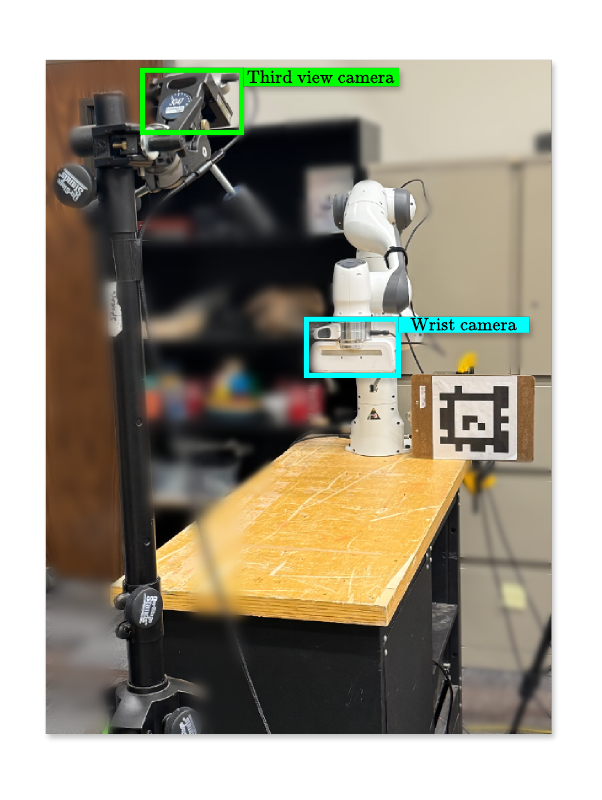}
    \vspace{-4em}
    \caption{Franka Emika Panda Arm Experiment Setup.}
    \label{fig:franka_setting}
    \vspace{-1em}
\end{figure}

\noindent \textbf{Hardware Setup.}
We conduct real-world tabletop manipulation experiments using a single Franka Emika Panda robotic arm and two RealSense D435 cameras, one mounted in a third perspective and one mounted as a wrist camera (\cref{fig:franka_setting}). The cameras capture images in 1920 × 1080 resolution, which is resized and center-cropped to construct 224 × 224 resolution inputs for training and inference.  

\begin{figure}[!h]
    \centering
    \includegraphics[width=\linewidth]{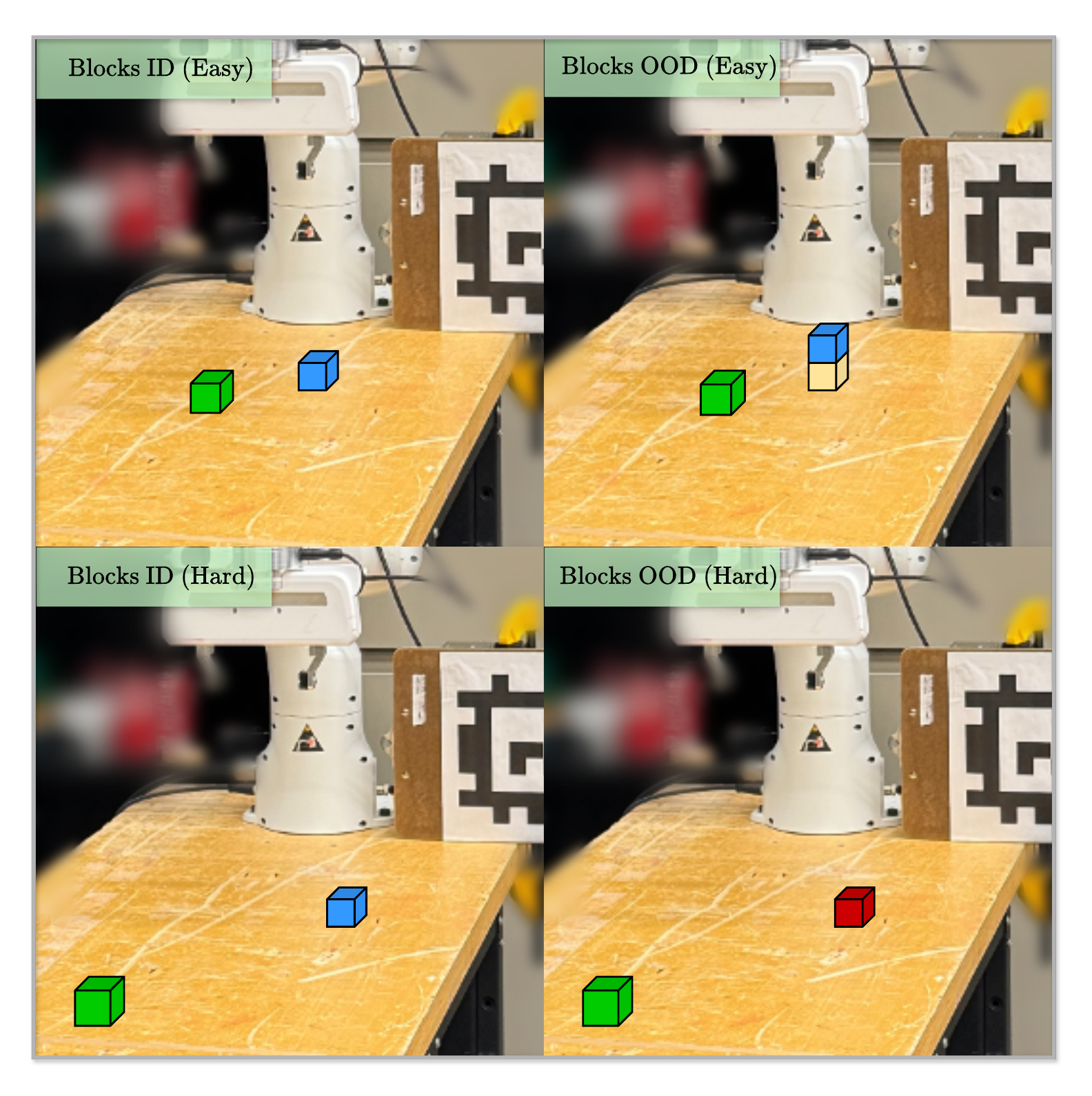}
    \caption{\textbf{Task 2 (Block) settings.} To further differentiate between MAPS and the baseline MiniVLA-OFT, we construct a more challenging "Hard" variant of T2: Block by introducing more geometric and vision variations. Blocks ID Hard (bottom left) increases the separation between the two blocks to probe long range planning and task execution, whereas Blocks OOD Hard (bottom right) additionally introduces an unseen new color block to the task.}
    \vspace{-2em}
    \label{fig:blocks-comparison}
\end{figure}

\noindent \textbf{Experiment Setup.}
We design 4 task settings:
\begin{enumerate}
    \item Coke: grab the coke can and lift it up,
    \item Blocks: stack the blue block on the green block,
    \item Cups: stack the orange cup onto the green cup,
    \item Laptop: close the laptop lid fully.
\end{enumerate}
We evaluate each task setting using ten rollouts across two object–location configurations, with five rollouts per configuration. For the multi-object Blocks and Cups tasks, the second configuration is created by swapping the positions of the objects. For the Coke task, we use “near’’ and “far’’ layouts, placing the can either within the gripper’s immediate reach or farther away. For the Laptop task, the two configurations place the laptop at an acute ~60° angle, oriented either toward or away from the camera. To enable a more fine-grained task analysis, we further introduce a Hard setting for the Blocks task, where objects are positioned farther from the robot arm (\cref{fig:blocks-comparison}) while still relying on configurations that simply swap object positions. We also define a new out-of-distribution (OOD) variant for this Hard setting by replacing the blue block with a red one. This results in a substantially more challenging OOD condition than the original Easy OOD, as both the visual appearance and semantic attributes differ from the pretrained setup.

\subsection{Qualitative Results}
\label{sec:franka_qualitative}

\begin{figure*}[!h]
    \centering

    \includegraphics[width=\linewidth]{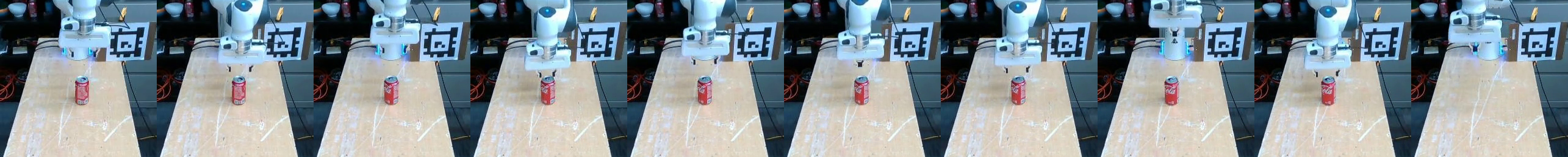}
    \includegraphics[width=\linewidth]{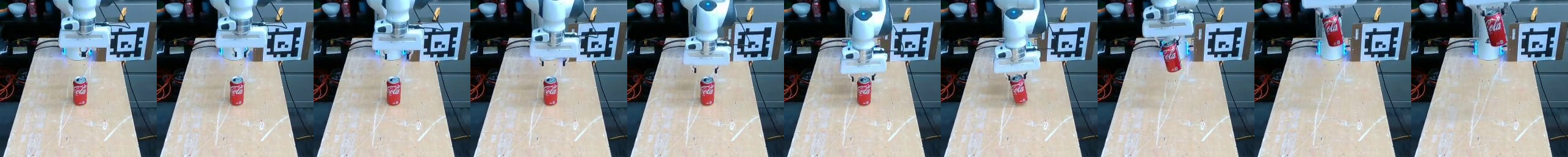}
    {\small {(1) Grab the Coke Can and Lift It Up}}\\[2mm]

    \includegraphics[width=\linewidth]{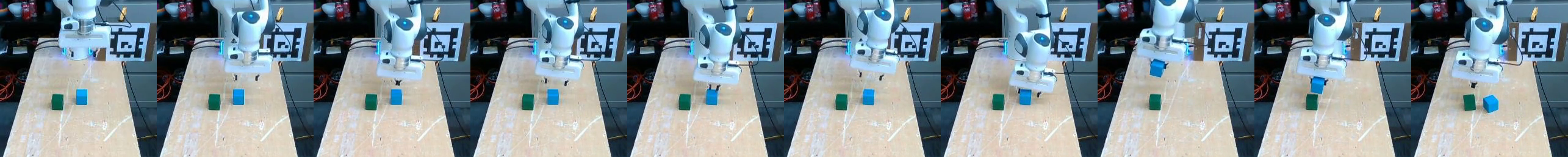}
    \includegraphics[width=\linewidth]{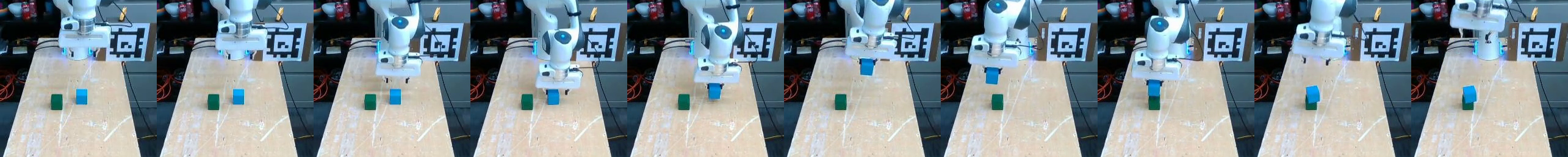}
    {\small {(2) Stack the Blue Block on the Green Block}}\\[2mm]

    \includegraphics[width=\linewidth]{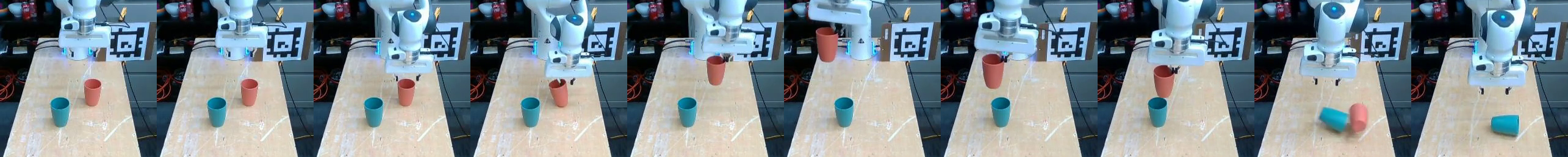}
    \includegraphics[width=\linewidth]{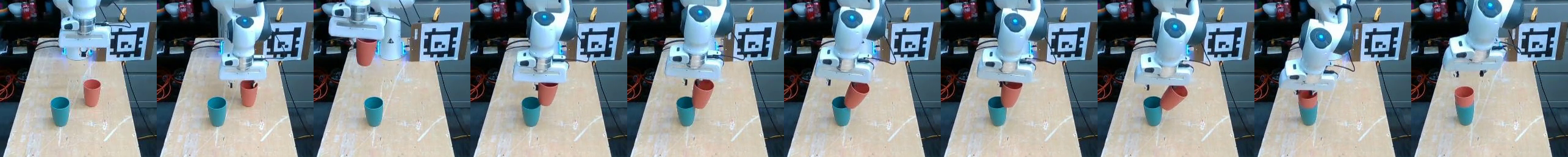}
    {\small {(3) Stack the Orange Cup onto the Green Cup}}\\[2mm]

    \includegraphics[width=\linewidth]{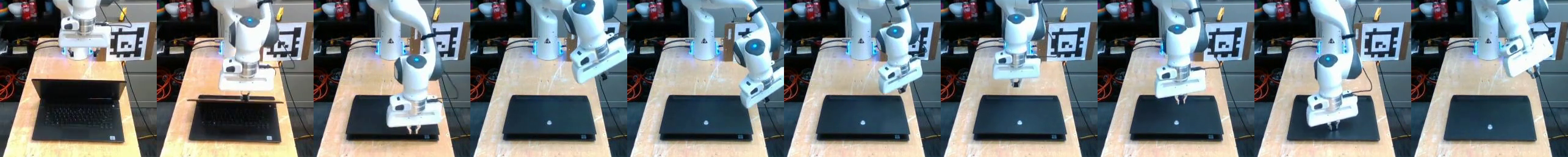}
    \includegraphics[width=\linewidth]{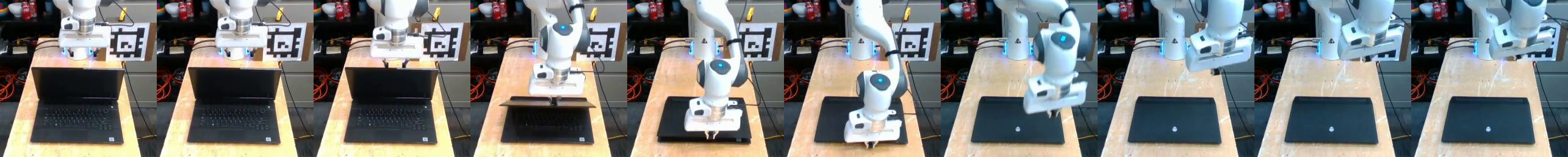}
    {\small {(4) Close the Laptop Lid Fully}}\\[2mm]

    \caption{ID tasks: baseline (top) vs. MAPS (bottom).}
    \label{fig:franka_id_video}
\end{figure*}

\begin{figure*}[!h]
    \centering

    \includegraphics[width=\linewidth]{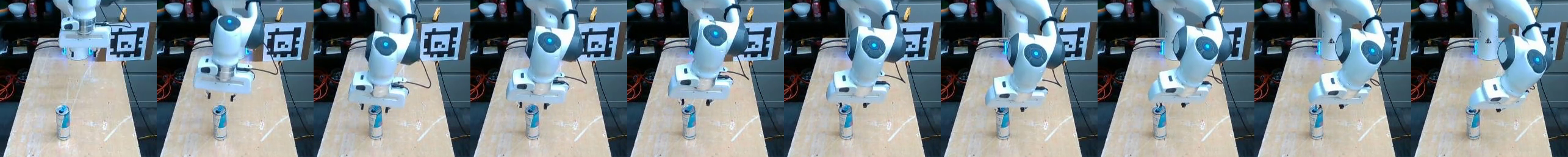}
    \includegraphics[width=\linewidth]{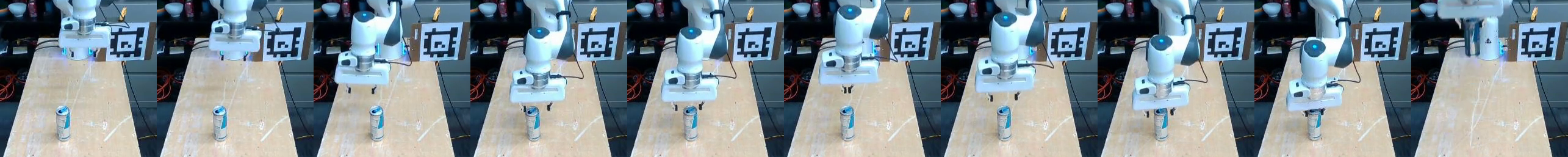}
    {\small {(5) Grab the Red Bull Can and Lift It Up}}\\[2mm]

    \includegraphics[width=\linewidth]{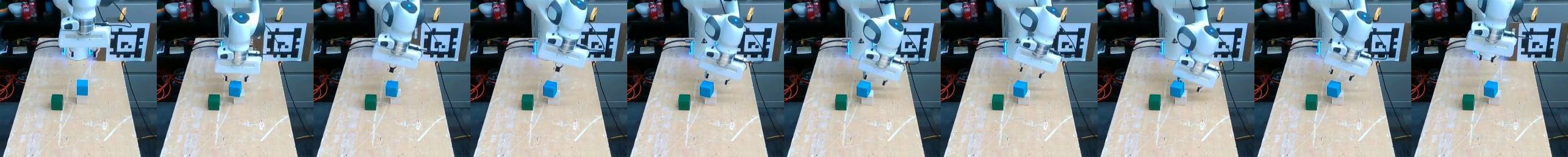}
    \includegraphics[width=\linewidth]{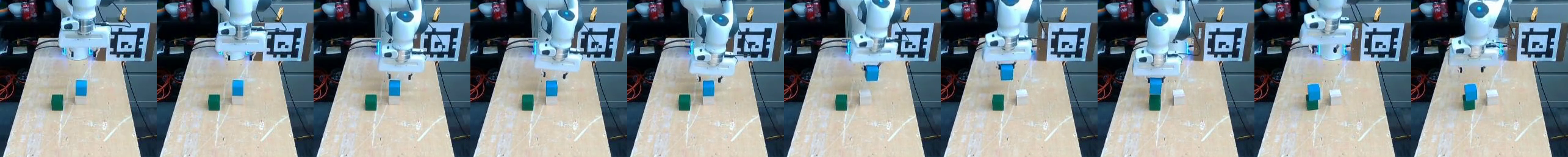}
    {\small {(6) Stack the Blue Block on the Green Block}}\\[2mm]

    \includegraphics[width=\linewidth]{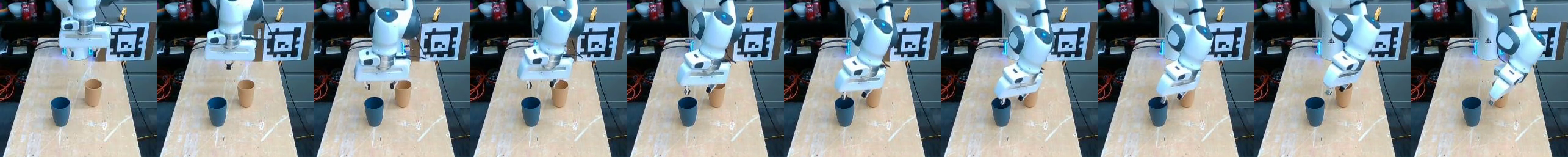}
    \includegraphics[width=\linewidth]{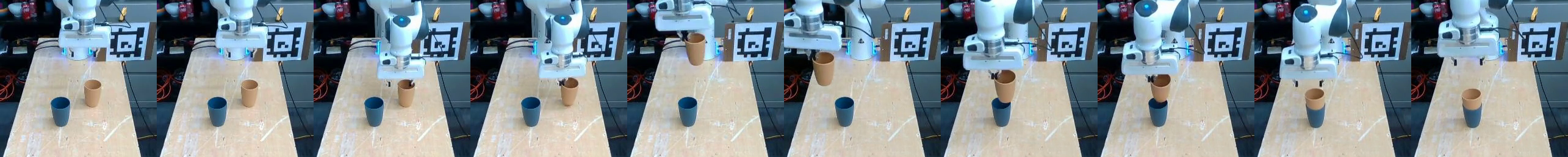}
    {\small {(7) Stack the Yellow Cup onto the Blue Cup}}\\[2mm]

    \includegraphics[width=\linewidth]{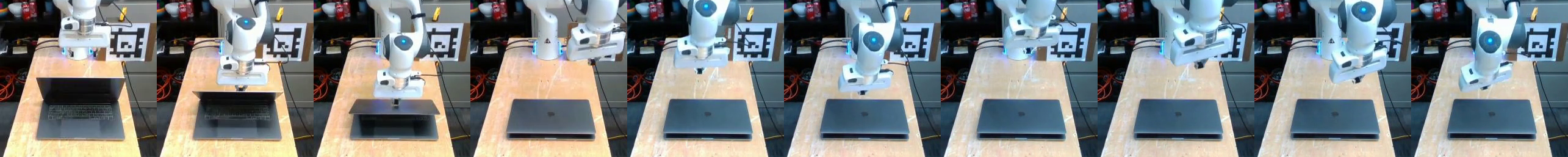}
    \includegraphics[width=\linewidth]{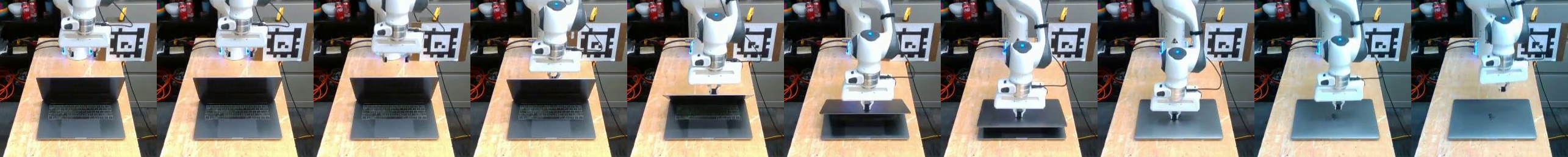}
    {\small {(8) Close the Laptop Lid Fully}}\\[2mm]

    \caption{OOD tasks: baseline (top) vs. MAPS (bottom).}
    \label{fig:franka_ood_video}
\end{figure*}

We show the qualitative results of ID and OOD tasks in \cref{fig:franka_id_video,fig:franka_ood_video}, with raw video recordings of these rollouts included on our project site. Across both settings, MAPS and the baseline exhibit clear and consistent behavioral differences. The baseline policies frequently stall, hesitate, or follow incorrect trajectories, resulting in missed grasps, poor approach angles, unstable placements, or premature object drops. These failures are visible in Cup ID, Block ID, and Coke ID (\cref{fig:franka_id_video}), where the baseline often grasps imprecisely, retries multiple times (Block ID), tips objects over (Coke ID), or performs poor localization that causes early releases (Cup ID). In contrast, MAPS executes smoother, more deliberate, and more stable motion plans. It firmly grasps objects, recalibrates effectively to locate secondary targets, and demonstrates careful multi-step execution -- such as reopening and reclosing the gripper when the orange cup initially fails to stack into the green cup -- without letting objects fall.

These qualitative differences become even more pronounced under distribution shifts (\cref{fig:franka_ood_video}). MAPS consistently adapts to altered object appearances and dynamics, whereas the baseline tends to rigidly reuse its ID trajectories and fails to compensate for changes. In Laptop OOD, replacing the Alienware laptop with a MacBook Pro introduces greater joint friction; the baseline follows its original trajectory and leaves the MacBook partially open (10–30\%), while MAPS adopts a more conservative path and applies sufficient, consistent force to close it fully. A similar pattern appears in Block OOD, where the first block’s elevation is altered. MAPS successfully adjusts its grasp strategy, while the baseline struggles and persists with its ID motion. Taken together, these visual rollouts show that MAPS not only corrects common baseline failure modes but also exhibits markedly stronger robustness, carefulness, and adaptability in both familiar and shifted environments.

\subsection{Results Discussion}
\label{sec:franka_discussion}
We discuss some questions regarding our qualitative and quantitative results.

\noindent\textbf{(1) Why are the performance gains on our real world ID evaluations so much more pronounced compared to our simulation ID results in the main paper?} One of our central hypotheses for the large ID performance gap on Franka is the limited size of our pretraining dataset -- only 600 demonstrations -- which likely leads to substantial overfitting in the baseline model. Both our qualitative observations and quantitative OOD results already indicate that the baseline MiniVLA-OFT is highly sensitive to spurious correlations. We believe the baseline has overfit so strongly to the demonstration distribution that minor natural variations which should remain well within the ID regime become challenging for it to handle. In contrast, the regularization introduced by MAPS mitigates this overfitting, enabling the model to generalize more robustly.

\noindent\textbf {(2) Why, in general, are these success rates the way that they are?} Across our tasks, a consistent trend emerges: grasping is challenging, and grasping round objects is especially difficult. In contrast, the Laptop task is comparatively trivial because it only requires the robot arm to push the object along a trajectory rather than perform a true grasp. The Block task is easier than Cup because the block’s square geometry provides stable grasping affordances. This explains why the baseline and MAPS perform similarly on Block grasping, whereas MAPS is substantially better on Cup grasping. As shown in \cref{fig:franka_fine}, the baseline’s primary failure point on Blocks (ID) is in locating the second object, while on Cups (ID) the main dropoff occurs at the very first grasp. The Coke task also involves grasping a cylindrical object, but it is more straightforward than Cup because only a single grasp is required. Taken together, these observations suggest the following expected difficulty ordering: Cup $>$ Block $>$ Coke $>$ Laptop. This is also consistent with our observed results.

\noindent\textbf{(3) Why does MAPS perform worse on Laptop ID compared to baseline MiniVLA-OFT?} On the Laptop ID/OOD tasks, we observed a recurring failure mode shared by both MAPS and the baseline, stemming partly from the task’s design. In configurations where the laptop screen faces the robot, once the arm pushes the lid down past a certain angle, the third-view camera can no longer reliably determine whether the laptop is fully closed. Because the learned policies typically lift the arm after pushing to avoid contacting the table, small partial closings also become indistinguishable from the wrist-view camera. The single MAPS ID failure arose from this ambiguity, and we observed similar issues in the OOD setting for both methods. Although one could debate how to interpret “success” for such borderline cases, we chose to evaluate results strictly and consistently. Overall, the Laptop task appears too saturated and simple to reveal meaningful differences in the ID setting -- making a 10\% change negligible -- whereas the 30\% gap in the OOD setting is far more consequential.

\section{Comparison with Baselines}
\label{sec:suppl_ablation}
\begin{table}[!h]
\centering
\caption{\textbf{Dual Encoders vs. MAPS} on LIBERO using MiniVLA-VQ.}
\resizebox{\linewidth}{!}{
\begin{tabular}{lccc}
\toprule
\textbf{Method} & \textbf{LIBERO-90 (ID)} & \textbf{Avg. OOD} & \textbf{Params} \\
\midrule
MiniVLA-VQ (baseline) \cite{wang2025vlaadaptereffectiveparadigmtinyscale} & 79 & 0.0 & 1.5B \\
+Dual-Encoder \cite{grover2025enhancinggeneralizationvisionlanguageactionmodels} & 75 & 2.75 & 1.5B \\
+Dual-Encoder (new DINOv2+SigLIP) & 79 & 3.75 &  2.5B\\
\rowcolor{gray!15}\textbf{+MAPS (Ours)} & \textbf{82} & \textbf{4.75} & 1.5B \\
\bottomrule
\end{tabular}}
\label{tab:abl_dual_revla_ftp_maps}
\end{table}

\noindent \textbf{Comparison with Dual Encoders.}
The comparison against different dual-encoder variants is presented in \cref{tab:abl_dual_revla_ftp_maps}. Notably, MAPS achieves the strongest performance among all compared methods while remaining the most compute-efficient. Dual-encoder approaches require instantiating a second trainable copy of the frozen vision encoder, which effectively doubles the parameter count of the vision stack. We implement this as the \textbf{Dual-Encoder (new DINOv2+SigLIP)} variant. A more efficient variant is proposed by \cite{grover2025enhancinggeneralizationvisionlanguageactionmodels}, in which one of the vision encoders is frozen while the other is kept trainable. We implement this as the \textbf{Dual-Encoder} variant. As shown in the table, neither dual-encoder variant matches the ID or OOD performance of MAPS. \textbf{Dual-Encoder} performs worst, and \textbf{Dual-Encoder (new DINOv2+SigLIP)} is slightly better but incurs substantially higher architectural cost. MAPS offers the best performance without additional cost.

\noindent \textbf{Comparison with Weight Interpolation.}
Weight interpolation methods such as \cite{dey2024revlarevertingvisualdomain} also introduces additional overhead: it depends on having a finetuned set of VLA weights available for interpolation, which means one training run to obtain those weights and another to perform the interpolation process. 

\noindent\textbf{Comparison with Bi-Level Optimization.} Bi-level optimization approaches that hyper-optimize the constraints between the fine-tuned and pre-trained weights, such as \cite{tian2023fasttrainableprojectionrobust}, incur roughly twice the training time. Under this setting, the resulting policy attains 79\% ID success and 4\% OOD success using VLA-Adapter, and 0\% ID and OOD sucess using MiniVLA-VQ, both of which fall short of MAPS while requiring significantly higher computational cost.

\section{Configurations for each Benchmark/Model}
\label{sec:config}
\begin{table*}[h!]
\centering
\caption{Configurations for Each Model and Benchmark.}
\label{tab:config}
\resizebox{\linewidth}{!}{
\begin{tabular}{lccccccccccc}
\toprule
\textbf{Benchmark} & \textbf{Model} & \textbf{lr} & \textbf{steps} & \textbf{batch size} & \textbf{use wrist} & \textbf{use proprio} & \textbf{use img aug} & \textbf{lora} & \textbf{GPUs} & \textbf{Training Time} & \textbf{$\lambda_{\max}$}\\
\midrule

\multirow{4}{*}{LIBERO} 
& MiniVLA-VQ  & 1e-5 & 50000 & 16*8  & no  & no & False & False & 8*A40 & 24h & 0.5 \\
& MiniVLA-OFT & 5e-5 & 50000 & 8*8   & yes & yes & True & False & 8*A40 & 19h & 3.2 \\
& VLA-Adapter & 1e-4 & 50000 & 8*8   & yes & yes & True & False & 8*A40 & 22h & 0.5 \\
& OpenVLA-OFT & 5e-5 & 50000 & 4*16  & yes & yes & True & r=32 & 16*A40 & 28h & 1.0 \\

\midrule
\multirow{2}{*}{CALVIN} 
& MiniVLA-OFT & 5e-5 & 40000 & 8*8   & yes & yes & True & False & 8*A40 & 19h & 2.5 \\
& OpenVLA-OFT & 1e-4 & 35000 & 4*16  & yes & yes & True & r=32 & 16*A40 & 28h & 1.5 \\

\midrule
\multirow{3}{*}{SimplerEnv} 
& MiniVLA-VQ  & 1e-5 & 50000 & 32*4 & no & no & False & False & 4*H200 & 10h & 0.5 \\
& MiniVLA-OFT & 5e-5 & 100000 & 8*8 & no & no & True & False & 8*A40 & 15h & 3.0 \\
& OpenVLA-OFT & 2e-4 & 100000 & 16*4 & no & no & True & r=32 & 4*H200 & 21h & 0.8\\

\midrule
Franka & MiniVLA-OFT & 5e-5 & 100000 & 8*8 & yes & no & True & False & 8*A40 & 14h & 1.5 \\ 

\bottomrule
\end{tabular}
}
\end{table*}

In \cref{tab:config}, we summarize the training configurations used across different models and benchmarks. Specifically, we report the optimal learning rate, total training steps, global batch size, usage of the wrist camera and proprioceptive state, whether image augmentation is applied, whether LoRA is used (and, if so, the chosen rank), the GPU setup, total training time, and the value of $\lambda_{\max}$ used in MAPS. 

\section{Algorithm}
\begin{algorithm}
\textbf{Given:} ordered layer set 
$\mathcal{L} = (\text{DINOv2}, \text{SigLIP}, \text{Bridge}, \text{Language})$ \\[0.5ex]
\textbf{Hyperparameters:} $\alpha, \beta_1, \beta_2, \epsilon, \lambda_{\max}$ \\[0.5ex]

\textbf{Pre-computation:} For $k = 1,\dots,|\mathcal{L}|$ define
\[
\lambda_k \coloneqq \lambda_{\max}\!\left( 1 - \frac{k-1}{|\mathcal{L}|-1} \right),
\]
so that $\lambda_1 = \lambda_{\max}$ (DINOv2) and $\lambda_{|\mathcal{L}|} = 0$ (Language). \\[1ex]

\textbf{Initialization:} 
$m_0 \leftarrow 0$, \ 
$v_0 \leftarrow 0$, \ 
$t \leftarrow 0$, \ 
$c_0 \leftarrow 0$ \\[1ex]

\textbf{For} $k = 1,\dots,|\mathcal{L}|$ \textbf{(from DINOv2 to Language)} \textbf{do} \\
\hspace{4mm}\textbf{while} $\theta_t$ not converged \textbf{do} \\
\hspace{8mm} $t \leftarrow t + 1$ \\
\hspace{8mm} $g_t \leftarrow \nabla_{\theta}\,\widetilde{\mathcal{L}}(\theta_{t-1})$ \\
\hspace{8mm} $m_t \leftarrow \beta_1 m_{t-1} + (1 - \beta_1)\, g_t$ \\
\hspace{8mm} $v_t \leftarrow \beta_2 v_{t-1} + (1 - \beta_2)\, g_t^2$ \\[0.5ex]
\hspace{8mm} \textbf{Bias correction:} \\
\hspace{12mm} $\widehat{m_t} \leftarrow \dfrac{m_t}{1-\beta_1^t}$,\quad
               $\widehat{v_t} \leftarrow \dfrac{v_t}{1-\beta_2^t}$ \\[0.5ex]
\hspace{8mm} \textbf{Adam update:} \\
\hspace{12mm} $\widetilde{\theta}_t \leftarrow  
\theta_{t-1} - \dfrac{\alpha \widehat{m_t}}{\sqrt{\widehat{v_t}} + \epsilon}$ \\[0.5ex]
\hspace{8mm} $c_t \leftarrow -g_t^\intercal(\theta_{t-1} - \theta_0)$ \\[0.5ex]
\hspace{8mm} \textbf{If} $c_t < 0$ \textbf{then} \\
\hspace{12mm} $\theta_t \leftarrow 
\widetilde{\theta}_t - \lambda_k\, r_t\bigl(\widetilde{\theta}_t - \theta_0\bigr)$ \\
\hspace{8mm} \textbf{else} \\
\hspace{12mm} $\theta_t \leftarrow \widetilde{\theta}_t$ \\
\hspace{8mm} \textbf{end if} \\
\hspace{4mm}\textbf{end while} \\
\textbf{end for}
\caption{Adam with MAPS.}
\label{algo:adam_maps}
\end{algorithm}
The full algorithm of MAPS is shown in 
\cref{algo:adam_maps}. The algorithm extends standard Adam optimization with a progressive Module-Wise Proximity Scheduling regularization strategy applied across an ordered stack of model layers. Beginning with the lowest-level representation (DINOv2) and proceeding sequentially toward the highest-level language layers, the method assigns each layer a monotonically decreasing proximal weight $\lambda_k$, with the strongest regularization applied to early layers and no regularization applied to the final layer. For each layer, the algorithm performs standard Adam updates by computing stochastic gradients, maintaining exponential moving averages of first- and second-order moments, and applying bias correction before taking an adaptive step in parameter space. After each tentative Adam update $\widetilde{\theta_t}$, the method evaluates a MAPS consistency signal $c_t$, defined as the negative inner product between the current negative gradient $-g_t^T$ and the displacement from the initialization $\theta_{t-1}-\theta_0$. A negative value indicates movement that contradicts the expected descent direction relative to the original initialization and triggers a correction: the updated parameters are pulled back toward the initialization by an amount proportional to both the layer-specific proximal weight $\lambda_k$ and the deviation ratio $r_t$. If the consistency signal is non-negative, the Adam update is accepted unchanged. By iterating this procedure until convergence at each layer before progressing to the next, the algorithm preserves the stability of foundational layers and components (DINOv2 and SigLIP), encourages smooth and initialization-aligned updates in intermediate modules, and allows full flexibility in the final language components. 

\end{document}